\documentclass[sigconf,nonacm]{acmart}
\AtBeginDocument{%
  \providecommand\BibTeX{{%
    \normalfont B\kern-0.5em{\scshape i\kern-0.25em b}\kern-0.8em\TeX}}}

\setcopyright{acmcopyright}
\copyrightyear{2018}
\acmYear{2018}
\acmDOI{10.1145/1122445.1122456}

\acmConference[Woodstock '18]{Woodstock '18: ACM Symposium on Neural
  Gaze Detection}{June 03--05, 2018}{Woodstock, NY}
\acmBooktitle{Woodstock '18: ACM Symposium on Neural Gaze Detection,
  June 03--05, 2018, Woodstock, NY}
\acmPrice{15.00}
\acmISBN{978-1-4503-XXXX-X/18/06}

\usepackage{svg}
\usepackage{subfig}

\begin{document}

\title{The Effect of Model Compression on Fairness in Facial Expression Recognition}

\author{Samuil Stoychev}
\affiliation{%
  \institution{University of Cambridge}
  \city{Cambridge}
  \country{United Kingdom}}
\email{ss2719@cantab.ac.uk}

\author{Hatice Gunes}
\affiliation{%
  \institution{University of Cambridge}
  \city{Cambridge}
  \country{United Kingdom}}
\email{hatice.gunes@cl.cam.ac.uk}

\renewcommand{\shortauthors}{Samuil Stoychev and Hatice Gunes}

\begin{abstract}
Deep neural networks have proved hugely successful, achieving human-like performance on a variety of tasks. However, they are also computationally expensive, which has motivated the development of model compression techniques which reduce the resource consumption associated with deep learning models. Nevertheless, recent studies have suggested that model compression can have an adverse effect on algorithmic fairness, amplifying existing biases in machine learning models. With this project we aim to extend those studies to the context of facial expression recognition. To do that, we set up a neural network classifier to perform facial expression recognition and implement several model compression techniques on top of it. We then run experiments on two facial expression datasets, namely the Extended Cohn-Kanade Dataset (CK+DB) and the Real-World Affective Faces Database (RAF-DB), to examine the individual and combined effect that compression techniques have on the model size, accuracy and fairness. Our experimental results show that: (i) Compression and quantisation
achieve significant reduction in model size with minimal impact on overall accuracy for both CK+DB and RAF-DB; (ii) in terms of model accuracy, the
classifier trained and tested on RAF-DB seems more robust
to compression compared to the CK+ DB; (iii) for RAF-DB, the different compression strategies do not
seem to increase the gap in predictive performance across
the sensitive attributes of gender, race and
age which is in contrast with the results on the CK+DB, 
where compression seems to amplify existing biases for gender. We analyse the results and discuss the potential reasons for our findings.
\end{abstract}



\maketitle

\section{Introduction}

Recent years have seen \textit{deep neural networks} (DNNs) achieve state-of-the-art performance on a variety of problems including face recognition \cite{DeepFaceRecognition}, cancer detection \cite{DeepLearningForCancerDiagnostics}, natural language processing \cite{DeepLearningForCancerDiagnostics}, etc. Deep learning has proved particularly effective at extracting meaningful representations from raw data \cite{ExtractRepr}. 

However, as the predictive performance of deep neural networks has increased, so has the size of deep learning architectures: Modern DNNs can consist of hundreds of millions of parameters \cite{SizeOfDNNs}, making them slow to train and hard to store. Deep learning’s growing computational cost has made it hard to deploy deep learning models on resource-constrained devices (e.g. mobile phones, robots, microcontrollers) which often lack the storage, memory or processing power to support large DNNs \cite{MLOnResourceConstrainedDevices, TFlitePaper, QuantisationAndTraining}. The high resource consumption associated with deep learning models has also been problematic in the light of initiatives such as the “Green-AI” \cite{GreenAI, GreenAIArticle} movement advocating for a reduction in the carbon emissions and the environmental impact associated with artificial intelligence. 

This has given rise to the development of \textit{model compression} strategies, which aim to reduce the size of deep learning models. Examples of model compression techniques include \textit{pruning} \cite{ToPruneOrNotToPrune}, \textit{quantisation} \cite{QuantisationAndTraining}, \textit{weight clustering} \cite{DeepCompression}, etc. We provide a more detailed overview of the compression strategies considered in this project in Section \ref{modelcompressiontechniques}. 

However, a couple of recent studies have suggested that, by reducing the network capacity of the DNNs, model compression can amplify existing biases: Hooker et al. \cite{hooker2020characterising} demonstrate that pruning and post-training quantisation can amplify biases when classifying hair colour on CelebA. This issue is also raised in a study by Paganini \cite{paganini2020prune} who discusses the effect of pruning on algorithmic fairness and proposes a framework for fair model pruning. 

With this work, we aim to extend the aforementioned studies by Paganini and Hooker et al. to the context of affective computing. In particular, we consider the task of \textit{facial expression recognition} (FER) where the model has to classify expressions based on images of human faces. To this end, we train FER models using the CK+DB and the RAF-DB, and implement three compression strategies (pruning, weight clustering and post-training quantisation) on top of them. We then evaluate and compare the performance of the baseline models against the performance of the compressed models, and analyse the results to address three research questions: 

\begin{itemize}
    \item \textbf{RQ1: \textit{“How effective is model compression in the context of FER?”}} That is, can compression techniques achieve a considerable reduction in the model size, while preserving a high level of predictive accuracy? 
    \item \textbf{RQ2: \textit{“Do model compression techniques amplify biases?”}} Here we seek to verify the claims by Paganini and Hooker et al. across a wider variety of compression techniques and in the context of FER. 
    \item \textbf{RQ3: \textit{“Is the impact on fairness identical across different compression techniques?”}} We are interested to know whether all compression strategies amplify biases to the same extent. 
\end{itemize}

This study extends the previous works by Paganini and Hooker et al. in three directions: 

\begin{itemize}
    \item \textbf{Extending the problem to affective computing:} We consider the problem of compression’s effect on fairness in the context of affective computing and, in particular, facial expression recognition. By comparison, the study by Hooker et al. is based on classifying hair colour on CelebA \cite{celeba}, and the study by Paganini considers object recognition and digit classification tasks. 
    
    The topic of model compression is particularly relevant to the field of affective computing as affective computation is increasingly performed on constrained devices such as robots \cite{affective_robots} and mobile phones \cite{affective_mobile_phones}. 
    \item \textbf{Considering more model compression techniques:} Our study involves three compression strategies (pruning, weight clustering and post-training quantisation) – one more by Hooker et al. (who consider pruning and post-training quantisation), and two more by the study by Paganini, which focuses solely on pruning. 
    \item \textbf{Considering the combined effect of compression techniques:} In practice, compression techniques are often combined together to form so called \textit{“compression pipelines”} \cite{DeepCompression}. That is why, we also consider two combinations of compression strategies (pruning with quantisation and weight clustering with quantisation). The previous studies have only examined the individual behaviour of compression techniques. 
\end{itemize}


\section{Background}\label{background}

\subsection{Algorithmic Fairness}\label{algorithmic_fairness}

Nowadays, machine learning algorithms are used to inform or automate decision-making across various fields of high social importance. Machine learning approaches have been used for automating recruitment in large companies \cite{PersonalityAssessment}, assigning credit scores \cite{CreditRatingsPrediction, CreditCardDefault} and anticipating criminal activity \cite{CriminalJustice} to name a few. 

The increasing impact of machine learning on our society has highlighted the importance of \textit{algorithmic fairness}. An algorithm is considered to be fair if its behaviour is not improperly influenced by sensitive attributes (e.g., a person’s gender or race) \cite{FairnessInML}. 

Nevertheless, recent studies have exposed the propensity of algorithms to be unfair and exhibit dangerous \textit{biases}  , potentially \textit{“reinforcing the discriminatory practices in society”} \cite{Reinforcing}. For example, Amazon’s AI recruitment tool has been reported to favour male applicants over female applicants \cite{AmazonRecruitmentTool}. Apple’s credit score has also been shown to systematically disadvantage women \cite{AppleCreditAlgorithm}. A study by Joy Buolamwini has demonstrated that popular facial analysis services perform disproportionately poorly on dark-skinned females \cite{GenderShades}. 

The increasing awareness of algorithmic biases has given rise to multiple fairness initiatives such the Algorithmic Justice League \cite{algorithmic}, IBM’s AI Fairness 360 \cite{IBMFairness360} and Google’s ML Fairness \cite{GoogleMLFairness}. 
Research into fairness in facial expression recognition has also started gaining momentum. Xu et al. \cite{XuECCV20} compared three different bias-mitigation approaches, namely, a baseline, an attribute-aware, and a disentangled
approach on two well-known data sets, Real-World Affected
Faces-Database and CelebFaces Attributes. 
Cheong et al. in \cite{CheongKG21} provided an overview and techniques that can be used for achieving fairness in facial affect recognition. 

Despite the large body of research which has studied the problem, though, there is still no consensus in the scientific community on what the precise definition of fairness should be. Multiple definitions of fairness have been proposed but none of them is a “silver bullet” that fits all use cases. Instead, the “right” choice of a fairness metric often depends on the specific context in which the algorithm is used \cite{context}. 

For this work, and in the context of facial expression recognition, we adopt the fairness definition of \textit{overall accuracy equality} \cite{overall}. The definition is akin to the concepts of predictive parity \cite{predictiveParity} and disparate mistreatment \cite{disparateMistreatment}, and states that \textit{a fair algorithm should have the same predictive accuracy regardless of any underlying sensitive attributes}. 

To express this formally, assume we have a facial expression recognition model that aims to predict a subject’s true expression $Y$ by producing a prediction $\hat{Y}$. Let the subject’s gender be denoted by $G$ and be equal to either $m$ (male) or $f$ (female)\footnote{Scheuerman et al. \cite{gender} have noted that in the context of facial analysis, it is useful to differentiate \textit{gender appearance} (whether a subject appears feminine or masculine) from \textit{gender self-identification} (whether a subject identifies as male or female). Throughout this report, we use the terms ``male'' and ``female'' to refer only to gender appearance and make no assumptions about the gender self-identification of the subjects.}. In that case, we expect that a fair model would have the following property: 

\begin{equation}
    P(Y = \hat{Y} | G = m) = P(Y = \hat{Y} | G = f)
\end{equation}

That is, we expect that a fair FER model would classify the expressions of male and female subjects with the same accuracy. 

\begin{figure}
  \centering
  \includegraphics[width=0.65\columnwidth]{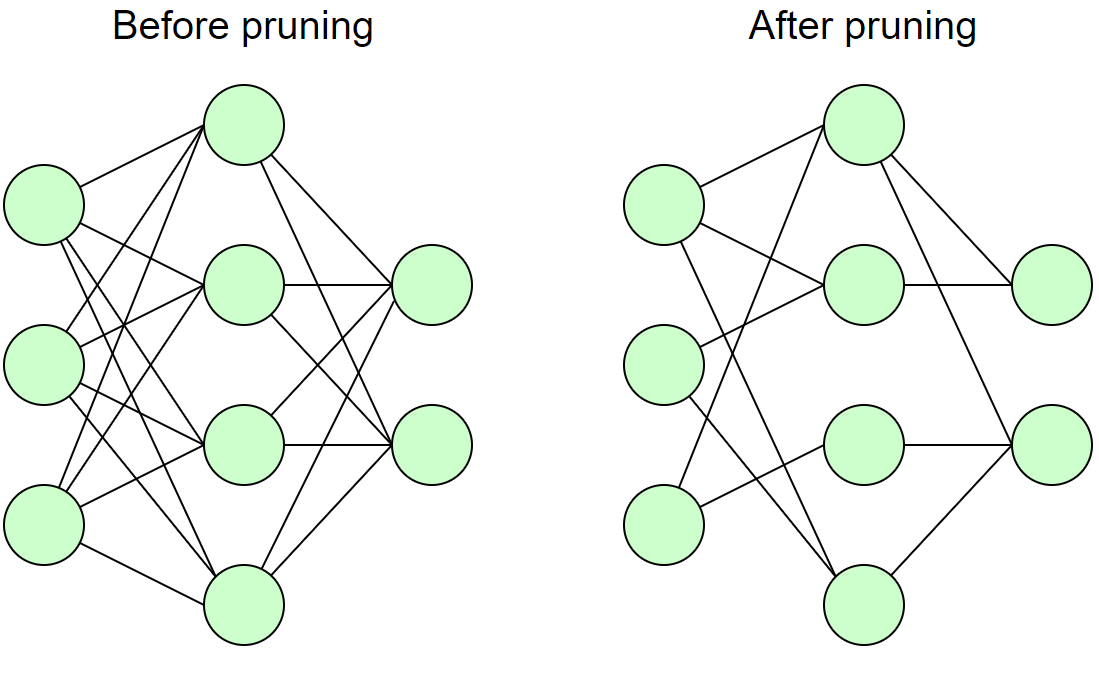}
  \caption{A DNN before and after pruning has been applied. The network has been stripped down to a subset of its original weights and the overall density has decreased. Illustration adapted from \cite{PruningDiagramSource}.}
  \label{fig:pruning}
\end{figure}

\subsection{Model Compression Techniques}\label{modelcompressiontechniques}

\subsubsection{Quantisation}

Quantisation is a popular compression method which can significantly reduce the size of the model, leading to savings in storage and memory \cite{SurveyOfCompressionAndAccelerationOfDNNs}. The key idea behind quantisation is sacrificing precision for efficiency – while most standard DNN implementations represent weights and activations using the \verb|float32| datatype, quantisation allows to represent those values using a smaller data type – normally \verb|float16| or \verb|int8| \cite{pruningplusquant}. Quantisation can either be introduced \textit{during} training (also known as \textit{quantisation-aware training}), or it can be applied to a pre-trained model, which is known as \textit{post-training quantisation} \cite{qaware}. In our experiments, we consider post-training quantisation to the \verb|int8| type. 

\subsubsection{Pruning}

Weight pruning is another compression strategy which can greatly reduce the size of a DNN \cite{pruningplusquant}. It does so by eliminating redundant weights which contribute little to the behaviour of the model. As illustrated in Figure \ref{fig:pruning}, pruning reduces the density of the neural network, making it more lightweight and easier to compress by traditional compression tools such as \verb|zip|\footnote{\url{https://www.tensorflow.org/model_optimization/guide/pruning}}. Which weights get pruned is dictated by the pruning strategy. The most popular approach, which we focus on in this project, is called \textit{magnitude-based pruning} \cite{StateOfPruning} and eliminates the weights with the lowest absolute value – i.e., the ones whose values are the closest to zero. The proportion of weights that need to be pruned is called the pruning \textit{sparsity} – for example, pruning at 90\% sparsity would remove 90\% of the connections in a given network.

\subsubsection{Weight Clustering}

\begin{figure}
  \centering
  \small
  \includegraphics[width=0.6\columnwidth]{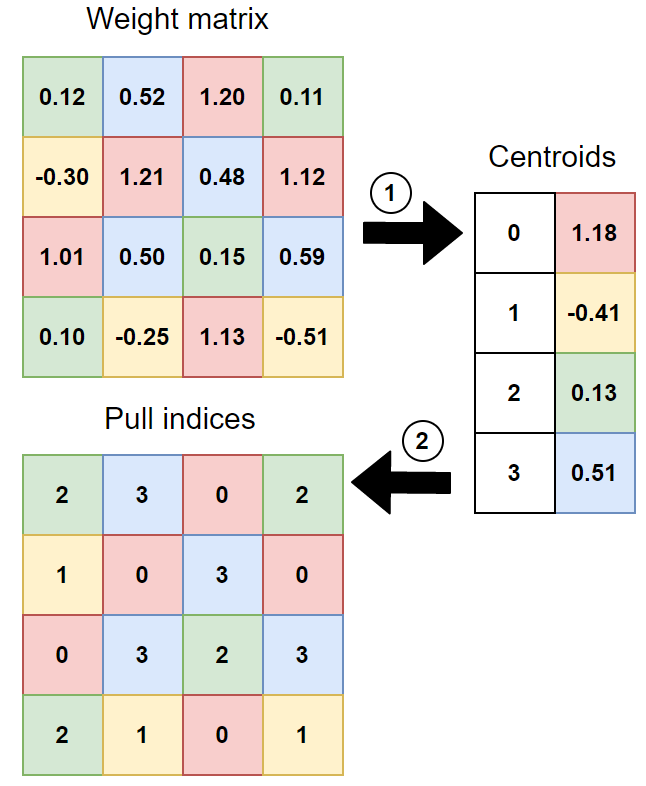}
  \caption{Diagram depicting the process of weight clustering. Similar weights get grouped together into the same cluster (step \textcircled{\raisebox{-0.9pt}{1}}), after which weights are replaced with pull indices (step \textcircled{\raisebox{-0.9pt}{2}}). Figure adapted from \cite{ClusteringDiagramSource}}
  \label{fig:clustering}
\end{figure}

Weight clustering (also known as \textit{weight sharing} \cite{DeepCompression}) reduces the size of the model by grouping together weight of similar values. This process is illustrated in Figure \ref{fig:clustering}: During step \textcircled{\raisebox{-0.9pt}{1}}, weight matrices are processed by a clustering algorithm, which maps each weight to one of $n$ clusters (where $n$ is the number of clusters specified by the user). Each cluster consists of an \textit{index} (one of $0, 1, ..., n-1$) and a \textit{centroid} value which is representative of the values of the weights mapped to that cluster. 
During step \textcircled{\raisebox{-0.9pt}{2}}, the weight matrices of the DNN are replaced with \textit{pull indices} – instead of containing the values of the corresponding weights, each pull index contains the index of the cluster containing the respective weight. During inference, the DNN model can use the pull indices to obtain the centroid values corresponding to each weight. 

Clustering reduces the model size for two reasons: First, float values only need to be stored to represent the $n$ centroid values. Meanwhile, the entire weight matrix is replaced with pull indices, each index represented by the smaller integer type. And second, the resulting pull indices are more likely to contain repeating values, making standard compression tools (e.g., \verb|zip|) more effective, similar to pruning. 

\begin{figure}
  \centering
  \subfloat[Train dataset.]{\includegraphics[width=0.8\columnwidth]{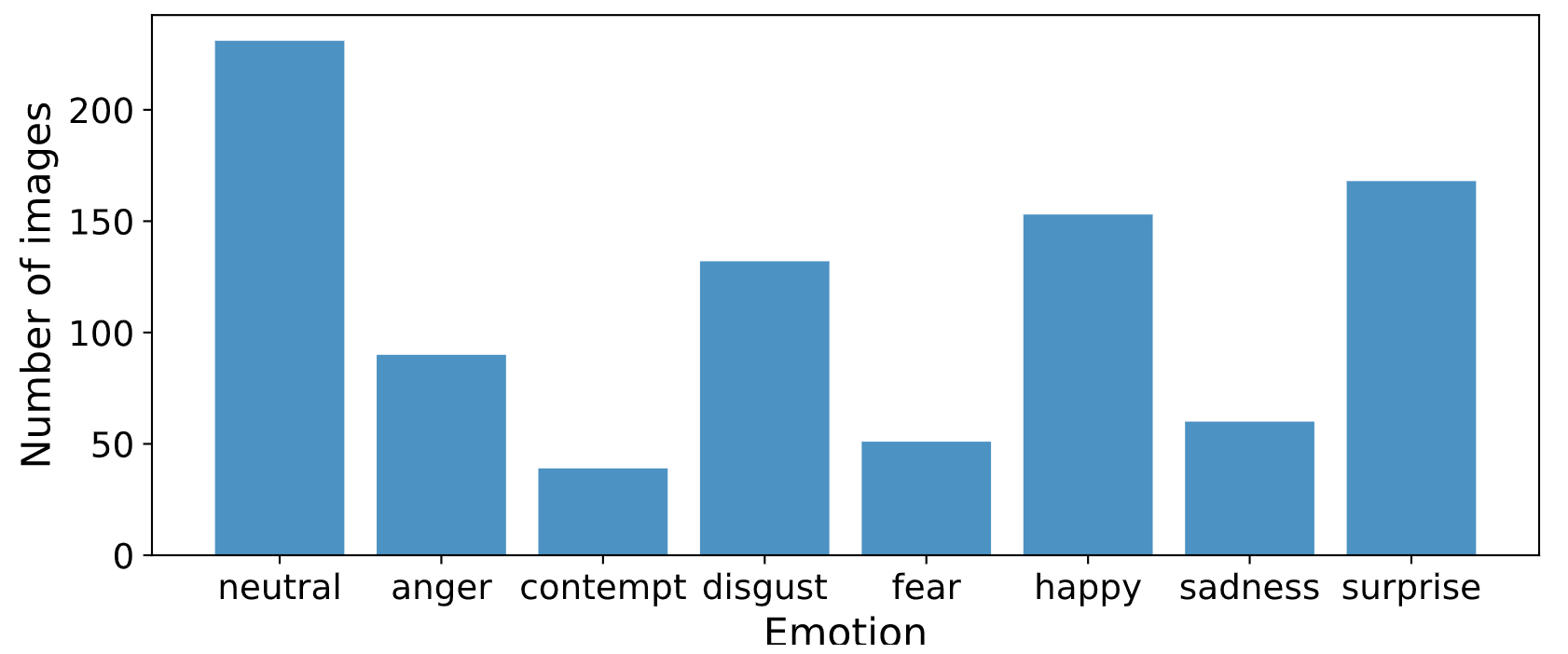}}
  \hfill
  \subfloat[Test dataset.]{\includegraphics[width=0.85\columnwidth]{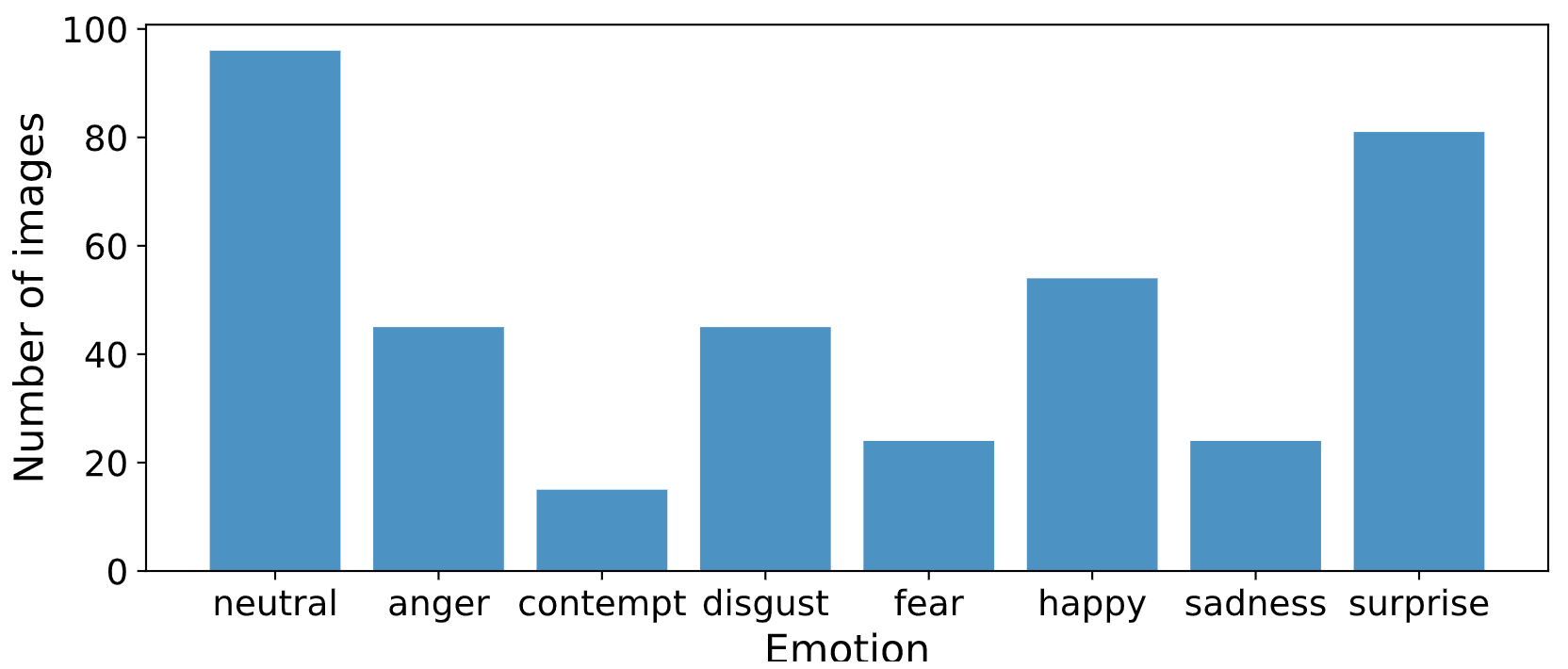}}
  \caption{Distribution of emotions across the train and test split of the CK+ dataset.}
  \label{fig:distributions}
\end{figure}

\section{Implementation}\label{implementation}

In this chapter, we summarise the implementation steps which the project has involved. A more detailed description of the technical implementation is available in the \verb|README.md| file of the code repository provided with the submission. 

\subsection{Data}\label{data_section}

To perform facial expression recognition, we make use of two popular datasets of human faces - CK+ \cite{ckplus} and RAF-DB \cite{rafdb}, which we describe below. 

\subsubsection{Extended Cohn-Kanade Dataset}

\begin{figure}
    \centering
    \includegraphics[width=0.5\textwidth]{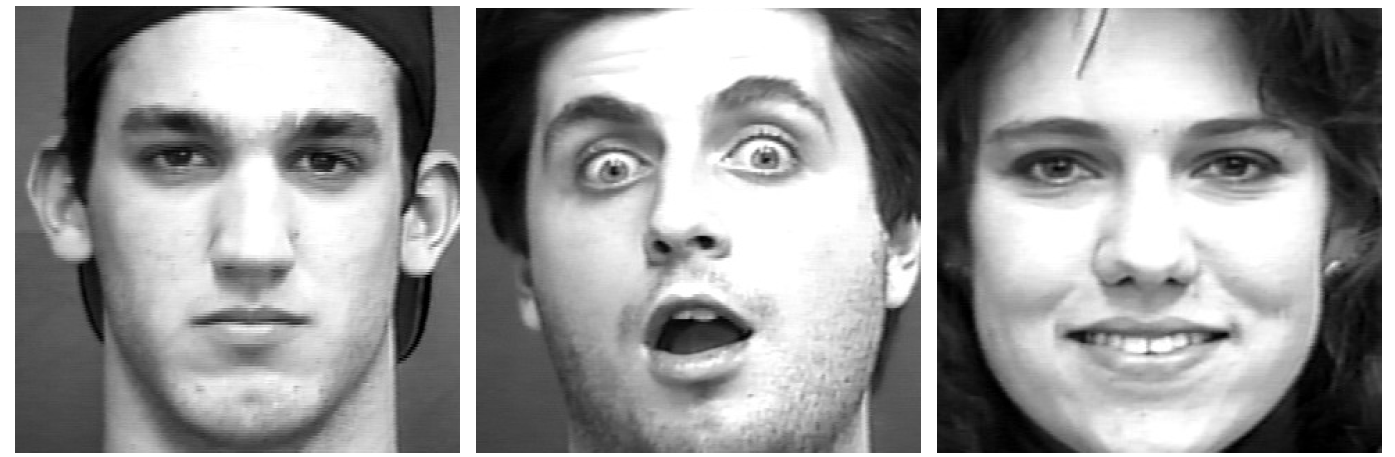}
    \caption{Images from the CK+ \cite{ckplus} dataset after cropping.}
    \label{fig:ckplus_samples}
\end{figure}

\begin{figure}
    \centering
 \includegraphics[width=0.5\textwidth]{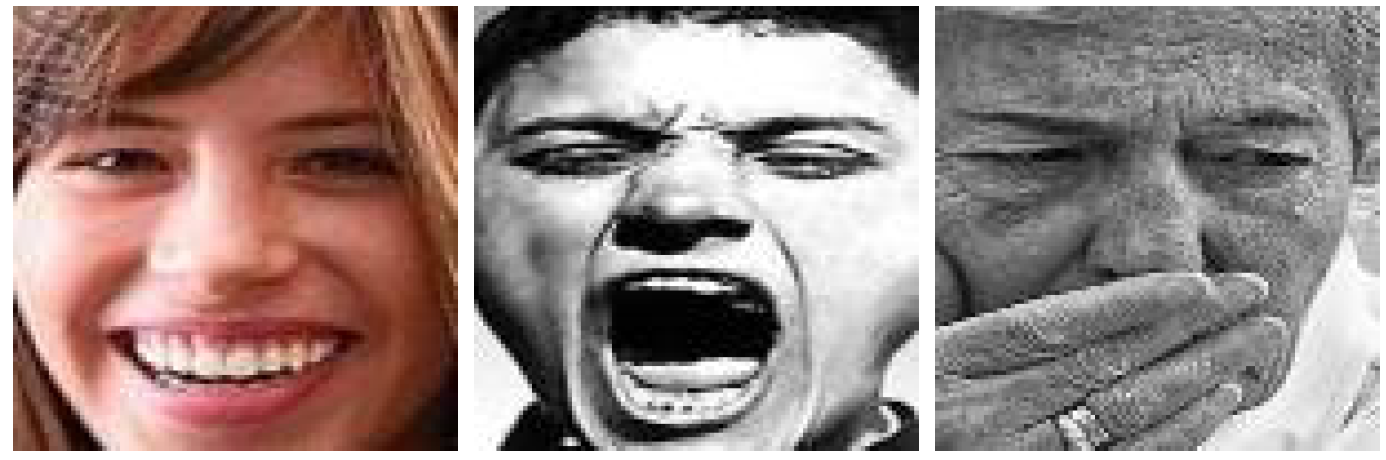}
    \caption{Images from the RAF-DB \cite{rafdb} dataset.}
    \label{fig:rafdb_samples}
\end{figure}

The Extended Cohn-Kanade Dataset (CK+) has been widely used in the context of facial expression recognition \cite{ckplus_usage1, ckplus_usage2}. It contains 327 labelled image sequences across 123 unique subjects, expressing one of 8 basic emotions - \textit{``neutral''}, \textit{``anger''}, \textit{``contempt''}, \textit{``disgust''}, \textit{``fear''}, \textit{``happy''}, \textit{``sadness''} and \textit{``surprise''}. Images have been obtained in a controlled lab environment with subjects facing the camera as illustrated in Figure \ref{fig:ckplus_samples}. 

In order to examine bias, we need annotations of demographic attributes. CK+ does not provide any annotations in that respect, so we manually annotate all 123 subjects based on their gender appearance. We assign a value \textit{“male”} if the subject looks masculine, and a value “\textit{female”} if they look feminine. The annotations are provided in the project repository under \verb|ckplus_labels.csv|. According to our annotation, the dataset consists of 84 female subjects and 39 male subjects. 

We then apply several pre-processing steps to the CK+ dataset: For each sequence in the dataset, we take the first frame to represent a neutral emotion, and the last 3 frames to represent the emotion which the sequence was annotated with (e.g. “happy”, “sad”, etc.). That is a common pre-processing step since CK+ sequences \textit{“are from the neutral face to the peak expression”} according to the dataset’s documentation. We then use the \verb|dlib|\footnote{\url{http://dlib.net/}} library to detect the faces of the subjects and crop the images around them (allowing an extra 10\% on each side to avoid cropping out parts of the chin, forehead or ears). 

For validation purposes, we split the original CK+ dataset into a \textit{train} and \textit{test} dataset. We use cross-subject validation, allocating 86 subjects to the train dataset and the other 37 to the test dataset. That gives us 924 images in total in the train dataset and 384 images in the test dataset. The train and test dataset follow a similar distribution with respect to the emotion labels as shown in Figure \ref{fig:distributions}. 

Finally, several data transformations are applied to the images using TensorFlow’s \verb|ImageDataGenerator|\footnote{\url{https://www.tensorflow.org/api_docs/python/tf/keras/preprocessing/image/ImageDataGenerator}}: Images are scaled down to $48 \times 48$ pixels and converted to grayscale (since CK+ contains some RGB images). Finally, to compensate for the relatively small size of the dataset, we augment the data by applying random horizontal flipping and random rotation in the range [$-10^{\circ}$, $10^{\circ}$]. 

\subsubsection{Real-World Affective Faces Database}

The Real-World Affective Faces Database (RAF-DB) \cite{rafdb} is another dataset of human faces which has been commonly used in the field of FER \cite{investigatingBiasAndFairness, largeScaleFER}. It provides 15,339 RGB images of human faces, aligned into squares of $100 \times 100$ pixels. Each image depicts one of seven basic emotions: \textit{``surprise''}, \textit{``fear''}, \textit{``disgust''}, \textit{``happiness''}, \textit{``sadness''}, \textit{``anger''} and \textit{``neutral''}. 

Unlike CK+ images, RAF-DB images are \textit{``in-the-wild''} - they have not been recorded in a controlled environment, so emotions are often expressed more subtly, lighting can vary and faces may be obfuscated as illustrated in Figure \ref{fig:rafdb_samples}. 

\begin{figure}
  \centering
  \subfloat[Train dataset.]{\includegraphics[width=0.8\columnwidth]{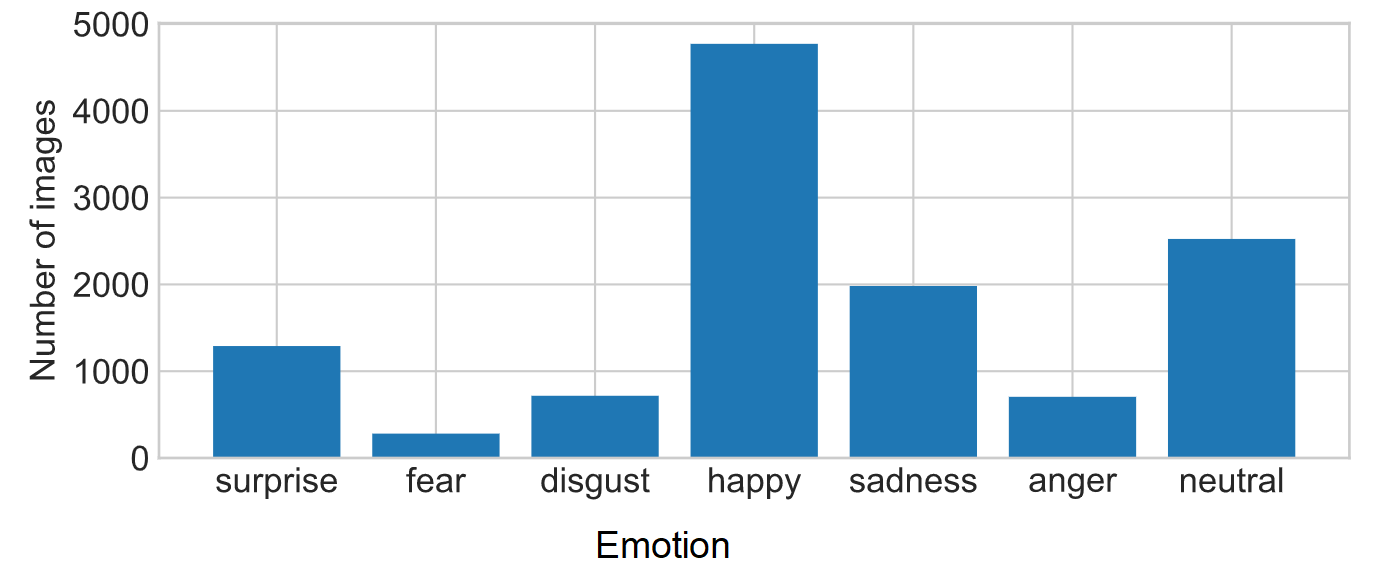}}
  \hfill
  \subfloat[Test dataset.]{\includegraphics[width=0.85\columnwidth]{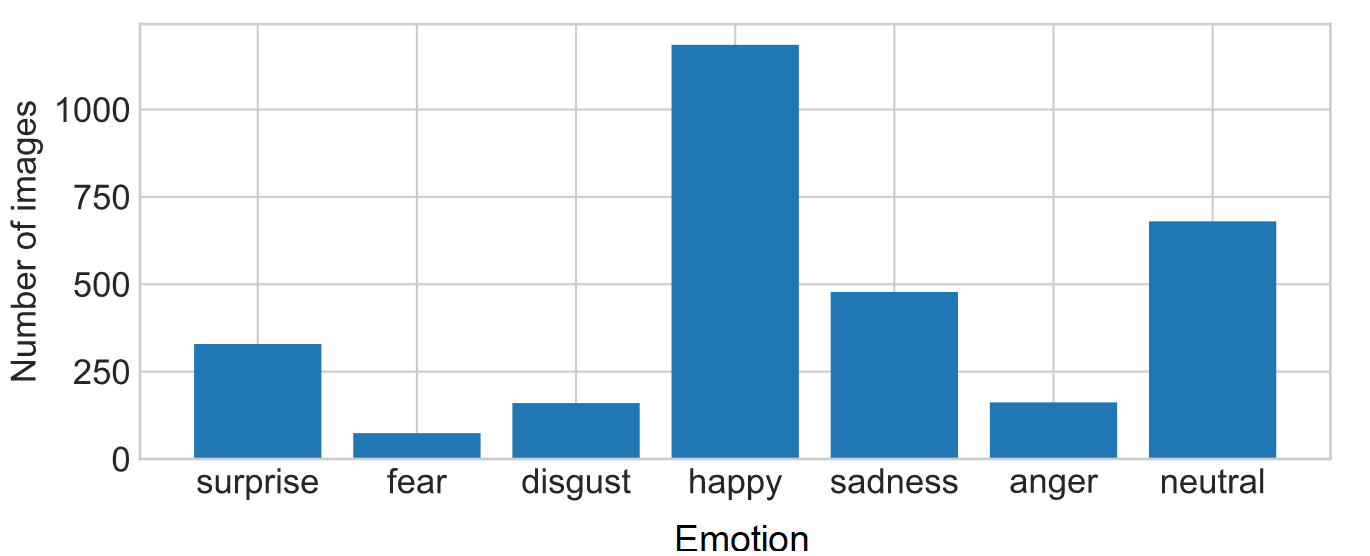}}
  \caption{Distribution of emotions across the train and test split of the RAF-DB dataset.}
  \label{fig:distributions_rafdb}
\end{figure}

Additionally, the RAF-DB dataset provides labels across three demographic categories - \textbf{gender} (with subjects labelled as one of \textit{``male''}, \textit{``female''} or \textit{``unsure''}), \textbf{race} (``Caucasian'', ``African-American'' or ``Asian'') and \textbf{age} (with subjects being assigned to one of 5 age groups - \textit{0-3}, \textit{4-19}, \textit{20-39}, \textit{40-69} and \textit{70+}). We form the train and test dataset preserving the original split defined by the authors \cite{rafdb}. This gives us 12271 training images and 3068 test images, with emotions distributed as shown in Figure \ref{fig:distributions_rafdb}.  

\subsection{Baseline Models}

\begin{figure}
  \centering
  \subfloat[Accuracy during training]{\includegraphics[width=0.50\columnwidth]{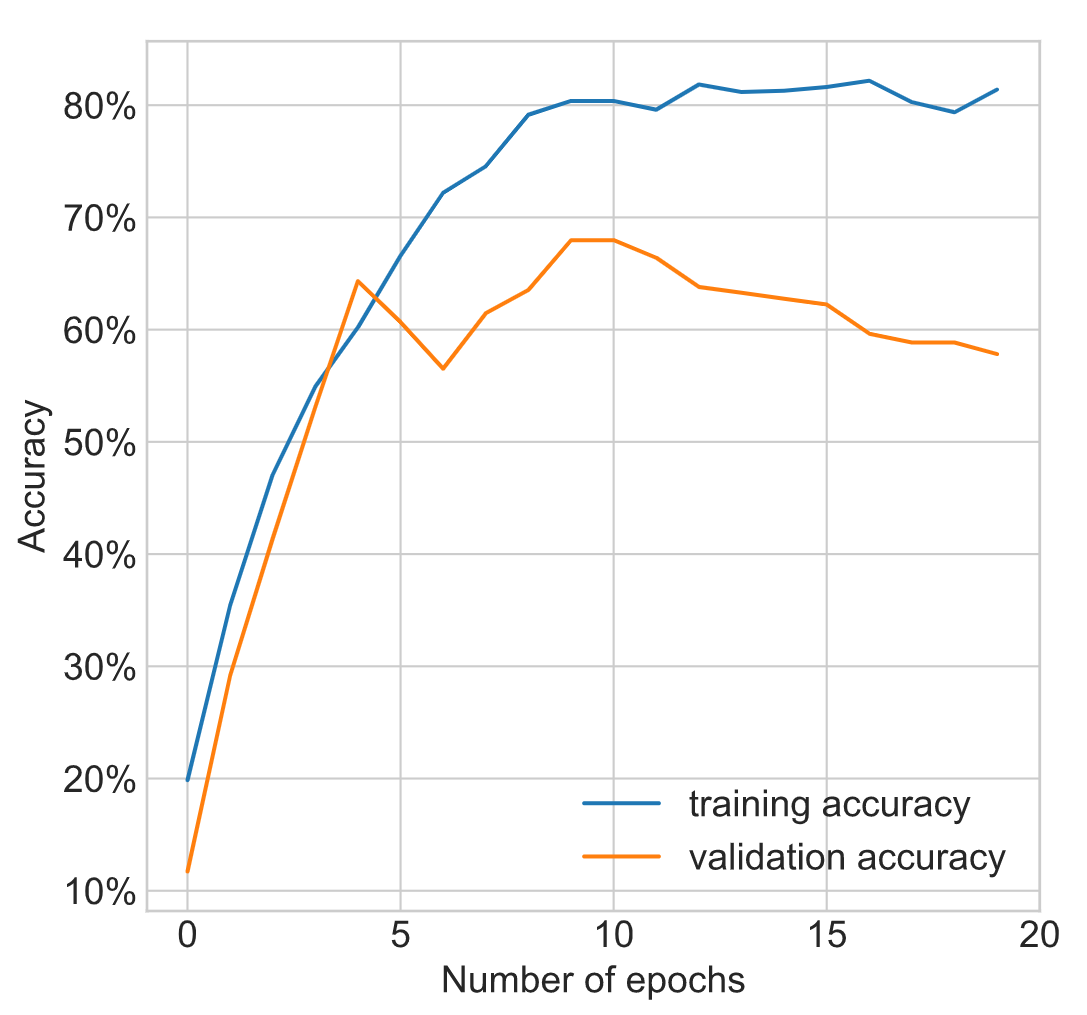}}
  \hfill
  \subfloat[Loss during training]{\includegraphics[width=0.50\columnwidth]{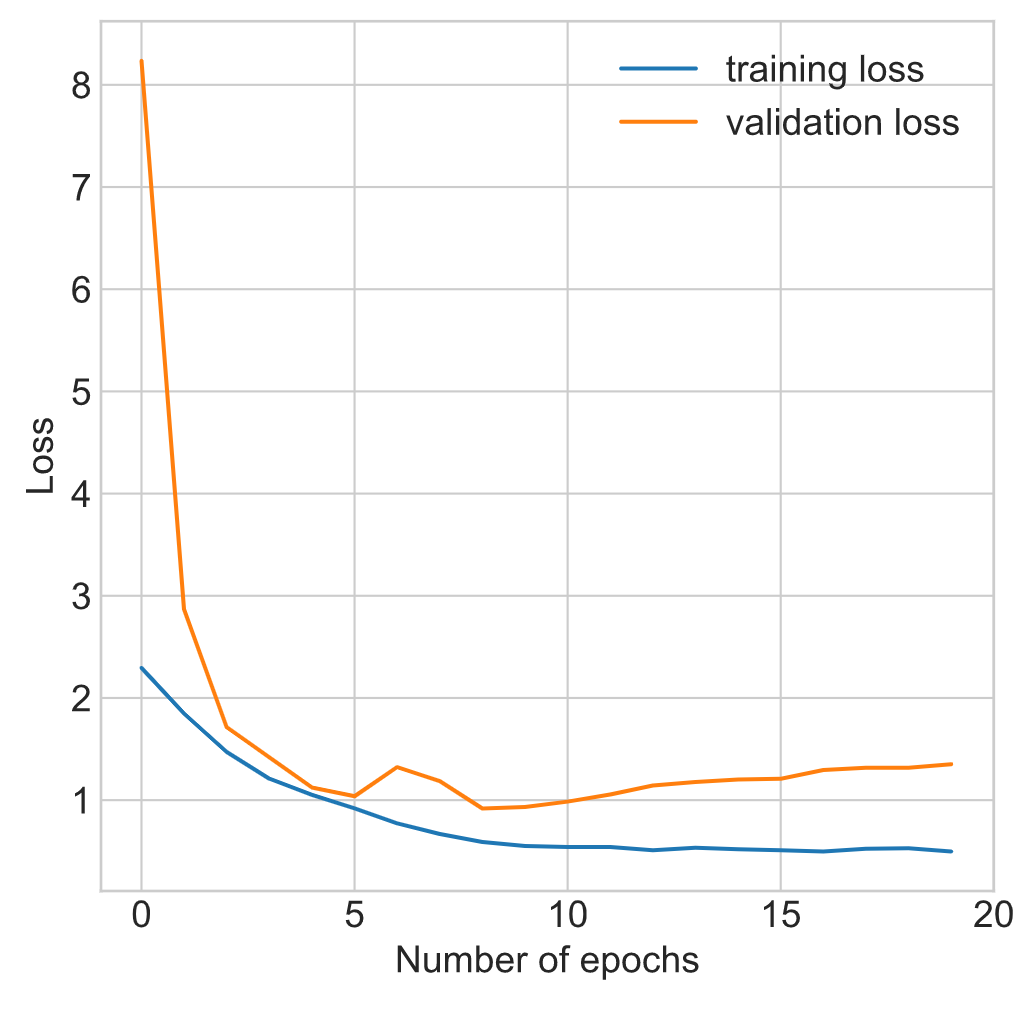}}
  \caption{Accuracy and loss during training the baseline CK+ classifier.} 
  \label{fig:training_process}
\end{figure}

\begin{figure}
  \centering
  \subfloat[Accuracy during training]{\includegraphics[width=0.50\columnwidth]{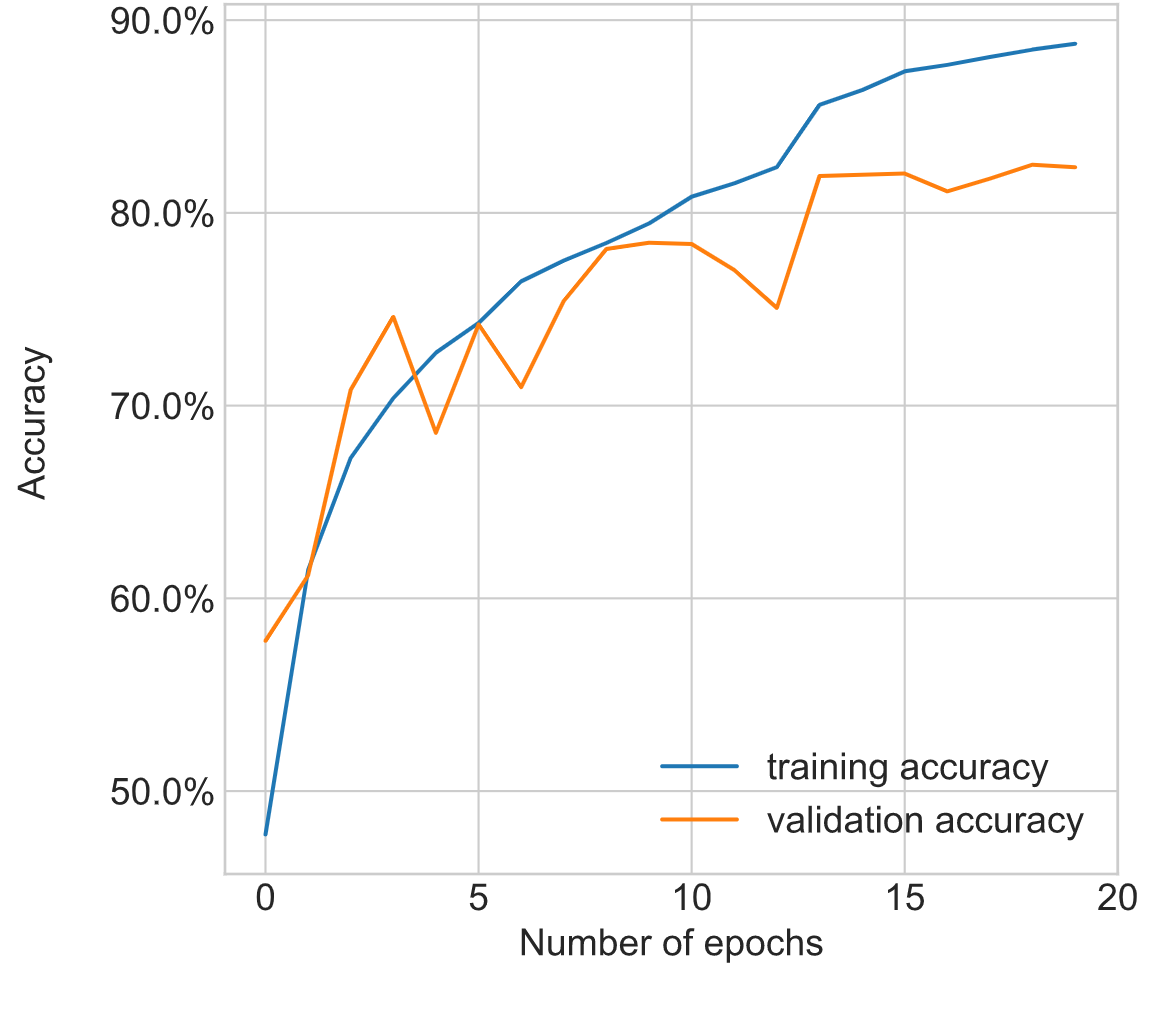}}
  \hfill
  \subfloat[Loss during training]{\includegraphics[width=0.50\columnwidth]{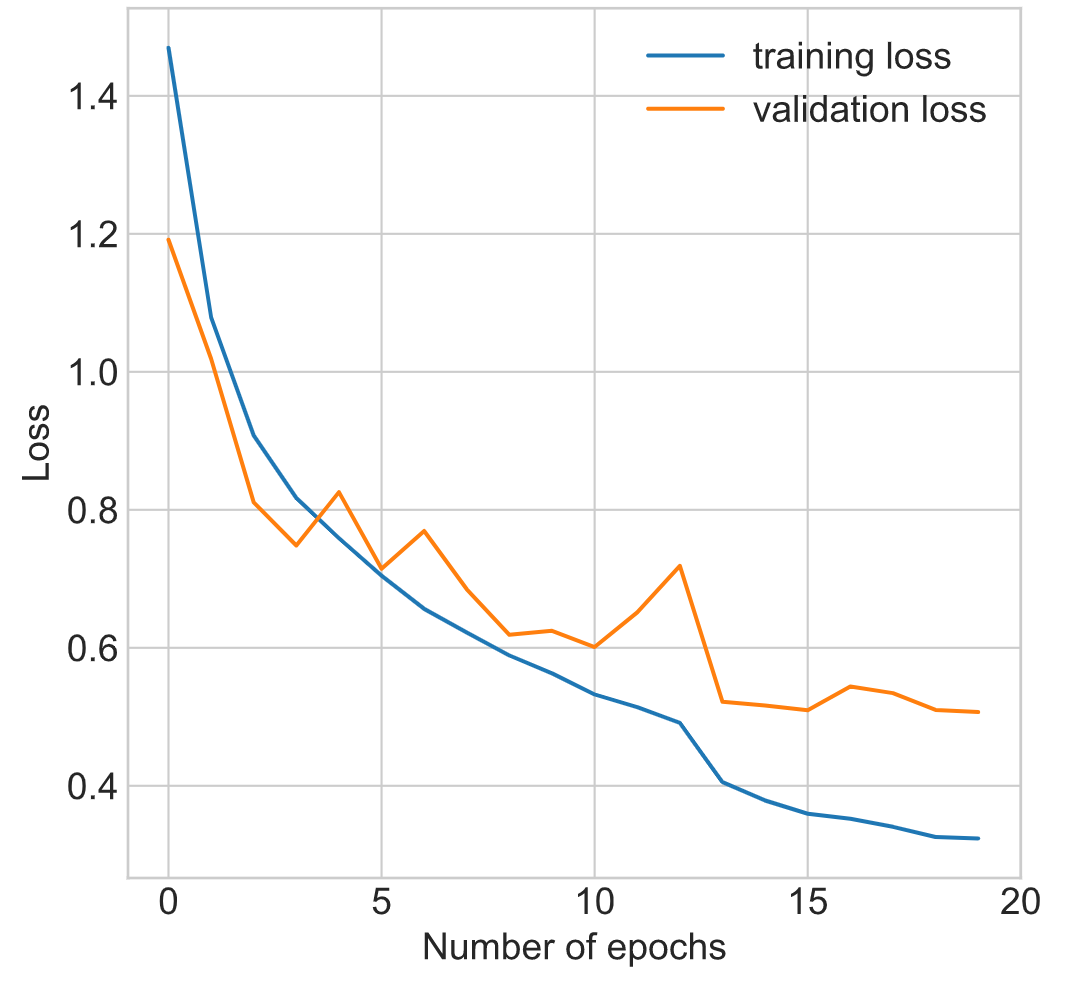}}
  \caption{Accuracy and loss during training the baseline RAF-DB classifier.}
  \label{fig:training_process_rafdb}
\end{figure}

We use the CK+ and RAF-DB datasets to implement two FER classifiers to serve as baselines to which we will apply the compression strategies. We follow an FER tutorial by S. Kekre \cite{FERTutorial} to set up a DNN classifier in Keras \cite{keras}. The architecture for the CK+ baseline is shown in Table \ref{tab:architecture}, and the RAF-DB architecture can be found in Table \ref{tab:architecture_rafdb} \-- the two are almost identical except for minor differences to account for the different input size and the different number of classes. The neural architecture is inspired by a study by Goodfellow et al. \cite{goodfellow2013challenges} and contains 4 convolutional layers, followed by 2 hidden fully-connected layers (with pooling and dropout layers in-between). At the time of its publication, the architecture achieved a then state-of-the-art performance of around 65\% on the FER-2013 dataset \cite{FERProof, goodfellow2013challenges}. 

We compile the baseline models using an Adam optimiser \cite{Adam} and a categorical cross-entropy loss \cite{CrossEntropy}. We train each model for 20 iterations keeping track of training and validation accuracy, and training and validation loss (where training is performed only on the train dataset and validation is performed only on the test dataset). At every iteration, we store the ``best'' weights observed so far (i.e., the ones associated with the highest validation accuracy). 

The process of obtaining the CK+ and the RAF-DB baseline is illustrated in Figure \ref{fig:training_process} and Figure \ref{fig:training_process_rafdb} respectively. The optimal weights are obtained after the $10^{\text{th}}$ iteration for CK+ and after the $19^{\text{th}}$ iteration for RAF-DB when the two models report respectively 67.96\% and 82.46\% validation accuracy. 


\begin{table}
  \caption{Summary of the architecture of the CK+ baseline model generated using Keras's \texttt{model.summary()}. The \texttt{None} values indicate the batch size is flexible.}
  \small
  \begin{tabular}{ccl}
    \toprule
    Layer & Output Shape\\
    \midrule
    \verb|conv2d| & \verb|(None, 48, 48, 64)|\\
    \verb|batch_normalization| & \verb|(None, 48, 48, 64)|\\
    \verb|activation| & \verb|(None, 48, 48, 64)|\\
    \verb|max_pooling2d| & \verb|(None, 24, 24, 64)|\\
    \verb|dropout| & \verb|(None, 24, 24, 64)|\\
    \verb|conv2d_1| & \verb|(None, 24, 24, 128)|\\
    \verb|batch_normalization_1| & \verb|(None, 24, 24, 128)|\\
    \verb|activation_1| & \verb|(None, 24, 24, 128)|\\
    \verb|max_pooling2d_1| & \verb|(None, 12, 12, 128)|\\ 
    \verb|dropout_1| & \verb|(None, 12, 12, 128)|\\
    \verb|conv2d_2| & \verb|(None, 12, 12, 512)|\\
    \verb|batch_normalization_2| & \verb|(None, 12, 12, 512)|\\
    \verb|activation_2| & \verb|(None, 12, 12, 512)|\\
    \verb|max_pooling2d_2| & \verb|(None, 6, 6, 512)|\\
    \verb|dropout_2| & \verb|(None, 6, 6, 512)|\\
    \verb|conv2d_3| & \verb|(None, 6, 6, 512)|\\
    \verb|batch_normalization_3| & \verb|(None, 6, 6, 512)|\\
    \verb|activation_3| & \verb|(None, 6, 6, 512)|\\
    \verb|max_pooling2d_3| & \verb|(None, 3, 3, 512)|\\
    \verb|dropout_3| & \verb|(None, 3, 3, 512)|\\
    \verb|flatten| & \verb|(None, 4608)|\\
    \verb|dense| & \verb|(None, 256)|\\
    \verb|batch_normalization_4| & \verb|(None, 256)|\\
    \verb|activation_4| & \verb|(None, 256)|\\
    \verb|dropout_4| & \verb|(None, 256)|\\
    \verb|dense_1| & \verb|(None, 512)|\\
    \verb|batch_normalization_5| & \verb|(None, 512)|\\
    \verb|activation_5| & \verb|(None, 512)|\\
    \verb|dropout_5| & \verb|(None, 512)|\\
    \verb|dense_2| & \verb|(None, 8)|\\ 
  \midrule
  Total trainable parameters: & 4,479,240\\
  \bottomrule
\end{tabular}
\label{tab:architecture}
\end{table}

\begin{table}[hb]
  \caption{Summary of the architecture of the RAF-DB baseline model generated using Keras's \texttt{model.summary()}. The \texttt{None} values indicate the batch size is flexible.}
  \small
  \begin{tabular}{ccl}
    \toprule
    Layer & Output Shape\\
    \midrule
    \verb|conv2d| & \verb|(None, 100, 100, 64)|\\
    \verb|batch_normalization| & \verb|(None, 100, 100, 64)|\\
    \verb|activation| & \verb|(None, 100, 100, 64)|\\
    \verb|max_pooling2d| & \verb|(None, 50, 50, 64)|\\
    \verb|dropout| & \verb|(None, 50, 50, 64)|\\
    \verb|conv2d_1| & \verb|(None, 50, 50, 128)|\\
    \verb|batch_normalization_1| & \verb|(None, 50, 50, 128)|\\
    \verb|activation_1| & \verb|(None, 50, 50, 128)|\\
    \verb|max_pooling2d_1| & \verb|(None, 25, 25, 128)|\\ 
    \verb|dropout_1| & \verb|(None, 25, 25, 128)|\\
    \verb|conv2d_2| & \verb|(None, 25, 25, 512)|\\
    \verb|batch_normalization_2| & \verb|(None, 25, 25, 512)|\\
    \verb|activation_2| & \verb|(None, 25, 25, 512)|\\
    \verb|max_pooling2d_2| & \verb|(None, 12, 12, 512)|\\
    \verb|dropout_2| & \verb|(None, 12, 12, 512)|\\
    \verb|conv2d_3| & \verb|(None, 12, 12, 512)|\\
    \verb|batch_normalization_3| & \verb|(None, 12, 12, 512)|\\
    \verb|activation_3| & \verb|(None, 12, 12, 512)|\\
    \verb|max_pooling2d_3| & \verb|(None, 6, 6, 512)|\\
    \verb|dropout_3| & \verb|(None, 6, 6, 512)|\\
    \verb|flatten| & \verb|(None, 18432)|\\
    \verb|dense| & \verb|(None, 256)|\\
    \verb|batch_normalization_4| & \verb|(None, 256)|\\
    \verb|activation_4| & \verb|(None, 256)|\\
    \verb|dropout_4| & \verb|(None, 256)|\\
    \verb|dense_1| & \verb|(None, 512)|\\
    \verb|batch_normalization_5| & \verb|(None, 512)|\\
    \verb|activation_5| & \verb|(None, 512)|\\
    \verb|dropout_5| & \verb|(None, 512)|\\
    \verb|dense_2| & \verb|(None, 7)|\\ 
  \midrule
  Total trainable parameters: & 8,013,703\\
  \bottomrule
\end{tabular}
\label{tab:architecture_rafdb}
\end{table}

\subsection{Model Compression Implementation}

We implement three model compression strategies – magnitude-based weight pruning, post-training quantisation and weight clustering. To do this, we make use of TensorFlow’s \textit{Model Optimization Toolkit}\footnote{\url{https://www.tensorflow.org/model_optimization/guide}}, part of the TFLite framework \cite{TFlitePaper}. 

\subsubsection{Quantisation} To quantise the model, we convert the pre-trained Keras baseline described in the last section to a TFLite model and apply the default TFLite optimisation strategy\footnote{\url{https://www.tensorflow.org/api_docs/python/tf/lite/Optimize}} which reduces the model representation to 8 bits. Finally, we store the quantised model on disk and compress  it using the \texttt{zip} compression tool, so that we are able to observe the change in size that quantisation has introduced. 

\subsubsection{Pruning} We apply pruning using TFLite’s \verb|ConstantSparsity|\footnote{\url{https://www.tensorflow.org/model_optimization/api_docs/python/tfmot/sparsity/keras/ConstantSparsity}} prnuing schedule. Our implementation is parameterised by the pruning sparsity – we observe the effect of this parameter on compression in Section \ref{experiments}. After pruning has been applied, we fine-tune the pruned model for 2 iterations as suggested by the TFLite documentation. Similarly to quantisation, we store the model on disk and compress it to evaluate the reduction in size. 

\subsubsection{Weight Clustering} We implement weight clustering using TFLite’s \verb|cluster_weights|\footnote{\url{https://www.tensorflow.org/model_optimization/api_docs/python/tfmot/clustering/keras/cluster_weights}} module and parameterise it by the number of clusters. Similar to pruning, we fine-tune the clustered model, store it on disk and compress it with \texttt{zip}. 

\subsubsection{Combined Compression} Additionally, our implementation allows applying quantisation on top of a pruned or a weight clustered model. This gives us two additional ``hybrid'' compression strategies - \textit{pruning with quantisation}, and \textit{weight clustering with quantisation}. We use those in our evaluation to explore the combined effect of compression techniques.

\section{Experiments}\label{experiments}

In this section we introduce the metrics we have considered during our experiments, and present the results we have obtained. We focus on the more interesting results from the study, however a more detailed breakdown of results is provided in the appendix as Table \ref{tab:extended_results} and Table \ref{tab:extended_results_rafdb}. 

\subsection{Metrics}

In our experiments, we compare the uncompressed baseline model against compressed versions of it across 3 metrics: 

\begin{itemize}
    \item \textbf{Model size} – that is the size of the model on disk in megabytes. This metric measures how effective a compression strategy is in reducing the storage requirement of the model. While some of the compression strategies could also reduce other system metrics such as latency, we focus on model size since all three compression techniques are primarily used to reduce storage consumption. 
    \item \textbf{Accuracy} – this is the overall accuracy a model achieves on the test dataset. We use this metric as an indicator of how compression has impacted predictive accuracy. 
    \item \textbf{Gender accuracy}  – to examine the fairness of the models, we also introduce the measure of gender accuracy, specifically as female vs. male accuracy. We define female accuracy as the number of correctly classified images containing a female subject over the total number of images containing a female subject. Similarly, male accuracy is the number of correctly classified images where the subject was male over the total number of images where the subject was male. Under the \textit{overall accuracy equality} formulation of fairness, which we defined in section \ref{algorithmic_fairness}, an unbiased model should have equal or similar female and male accuracy metrics. Conversely, a large discrepancy between the model’s accuracy for males and females would be a strong indicator of algorithmic bias. Just like overall accuracy, male and female accuracy are measured on the test dataset.
    \item \textbf{Race accuracy}  – this is similar to the gender accuracy, but this time we aim to examine the fairness of the models for different race groups as defined in the corresponding dataset. An unbiased model should have equal or similar accuracy for different race groups. Conversely, a large discrepancy between the model’s accuracy for different race groups would be a strong indicator of algorithmic bias. These are measured on the test dataset.
    \item \textbf{Age accuracy}  – this is similar to the gender and ethnicity accuracy, but this time we aim to examine the fairness of the models for different age groups as defined in the corresponding dataset. An unbiased model should have equal or similar accuracy for different age groups. These again are measured on the test dataset.
\end{itemize}

\subsection{CK+ Experiments}

\subsubsection{Baseline Performance}

We first report the baseline performance against which we compare the performance of the compressed models. As mentioned previously, the baseline model reports an overall accuracy of \textbf{67.96\%} on the test dataset. We measure the baseline’s size in the same way we measure the size of the compressed models – we save the model as a file on disk and apply \texttt{zip} compression to the file. After doing that, we find that the baseline’s model size on disk is 16,512,048 bytes or around \textbf{16.51 megabytes}. 

We find that the female accuracy of the baseline model is \textbf{67.08\%} and the male accuracy is \textbf{69.44\%}. This is interesting because as we mentioned in Section \ref{data_section}, the CK+ dataset is imbalanced in favour of female subjects and therefore we would expect the baseline model to classify females more accurately. 

One reason why the classifier might perform slightly better on male faces is the slight difference in the distribution of emotions across males and females in CK+ – for instance, only 2.1\% of male subjects have expressed “contempt” while female subjects have expressed this emotion more than twice more frequently (5.02\%). If contempt is an emotion that is inherently harder to classify, then this difference could translate into a minor advantage for classifying male subjects. In any case, though, the gap between male and female accuracy is too minor to conclude the baseline model is biased. 

\begin{figure}
  \centering
  \includegraphics[width=0.8\columnwidth]{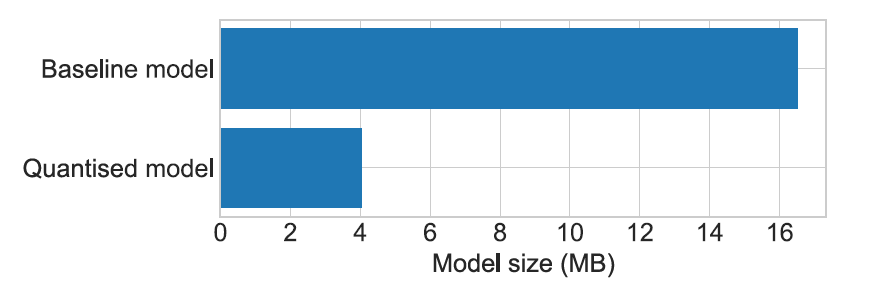}
  \caption{Size of the CK+ classifier before (top) and after (bottom) applying quantisation.}
  \label{fig:quantisation}
\end{figure}

\subsubsection{Quantisation Results}

We evaluate quantisation by quantising the baseline model 3 times and reporting the mean values for each metric. 

After applying quantisation to the baseline model, we observe a $4\times$ reduction in the model size as illustrated in Figure \ref{fig:quantisation}. Moreover, this compression comes at no cost – there is no change in the predictive accuracy or fairness whatsoever: Overall accuracy, female accuracy and male accuracy have all remained completely identical to those of the baseline model. Quantisation therefore preserves the fairness and predictive accuracy of the model, while introducing a significant reduction in its size. 

\subsubsection{Pruning Results}

We evaluate the pruning strategy at 6 different levels of sparsity: 10\%, 20\%, 30\%, 40\%, 50\% and 60\%. For each level of sparsity, we prune the baseline model three times and report the mean values for each metric. 

\begin{figure}
  \centering
  \subfloat[Model size after pruning.]{\includegraphics[width=0.8\columnwidth]{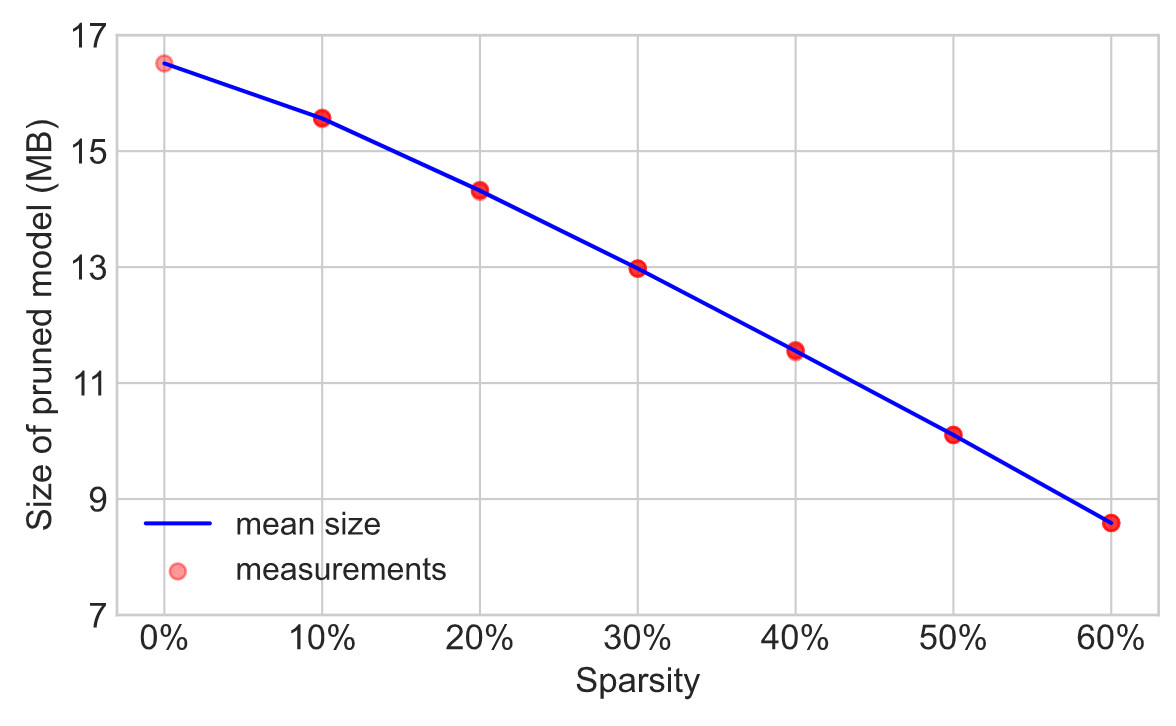}}
  \hfill
  \subfloat[Model accuracy after pruning. Standard deviation is shaded in gray.]{\includegraphics[width=0.8\columnwidth]{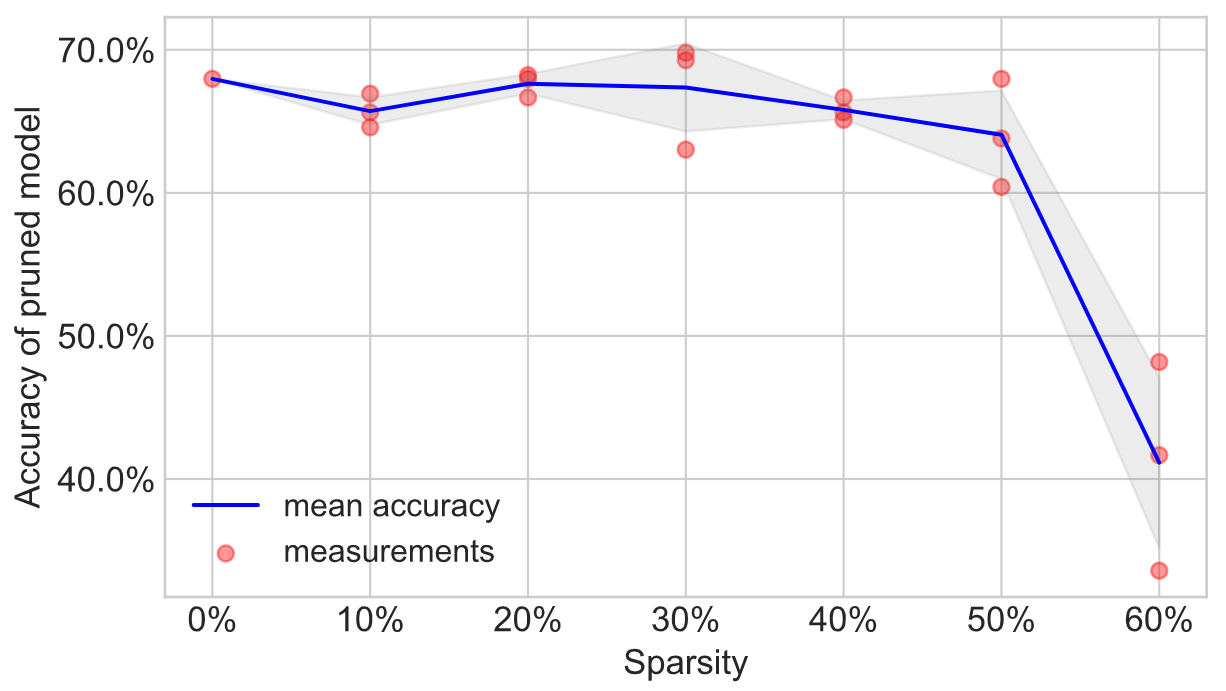}}
  \caption{Pruning's effect on model size and accuracy for CK+ DB.}
  \label{fig:pruning_tradeoff}
\end{figure}

We can observe the trade-off that pruning sparsity introduces in Figure \ref{fig:pruning_tradeoff}: As sparsity increases, the model size decreases linearly, reducing the model size by a half at 60\% sparsity. However, high sparsity also reduces the network capacity which can impact the accuracy of the model \cite{paganini2020prune}. We can see in Figure \ref{fig:pruning_tradeoff} (b) that overall accuracy steadily starts to decrease at 40\% and 50\% sparsity before plummeting at 60\%. 

Except for the considerable drop of accuracy at 60\% sparsity, though, we can conclude that pruning has had little impact on overall accuracy: At 50\% pruning, accuracy has only dropped from 67.96\% down to 64.06\%. 

However, despite the minor drop in overall accuracy, we find that pruning has dramatically increased the discrepancy between female and male accuracy: Table \ref{tab:pruning_fairness} shows that pruning the model tends to keep male accuracy high (in fact, male accuracy has increased for all sparsities up to 50\%) while deteriorating female accuracy (which has dropped by 7.22\% at 50\% sparsity).\footnote{Interestingly, pruning at 60\% actually reports better female than male accuracy. However, at 60\% sparsity both male and female accuracy are too low for such a model to be of practical interest.}

Those results are in agreement with the study by Hooker et al., which finds that \textit{“minimal changes to overall accuracy hide disproportionately high errors”} \cite{hooker2020characterising} in subgroups. Pruning the baseline model by up to 50\% has led to a considerable reduction in size with seemingly low impact on overall accuracy. However, while overall accuracy has stayed reasonably high, the initial 2.36\% gap between male and female accuracy has grown to 15.6\%, amplifying the minimal bias in the baseline model. 

\begin{table}
  \caption{Pruning's effect on fairness for CK+ DB.}
  \label{tab:pruning_fairness}
  \small
  \begin{tabular}{c|cc|c}
    \toprule
    Sparsity&Female accuracy&Male accuracy&Gap\\
    \midrule
    0\% (baseline) & 67.08\% & 69.44\% & 2.36\%\\
    10\% & 59.86\% & 75.46\% & \textbf{15.60}\%\\
    20\% & 62.77\% & 75.69\% & 12.92\%\\
    30\% & 62.36\% & 75.69\% & 13.33\%\\
    40\% & 63.47\% & 69.67\% & 6.20\%\\
    50\% & 59.86\% & 71.06\% & 11.20\%\\
    60\% & 44.02\% & 36.34\% & 7.68\%\\
  \bottomrule
\end{tabular}
\end{table}

As mentioned previously, we are also interested in the “combined” effect of compression techniques. To this end, we apply quantisation on top of the pruned models. We find that quantisation can greatly enhance the compression effect of pruning: Quantising the pruned models has decreased their size by a further 3.5 times on average (exact results reported in Table \ref{tab:extended_results} in the appendix).

Meanwhile, quantisation has not changed the overall accuracy, male accuracy or female accuracy of the pruned models by more than 1\%. Again, applying quantisation is “for free” since no significant impact on predictive performance or fairness is observed. 

\subsubsection{Weight Clustering Results}

We evaluate weight clustering with 4, 8, 16, 32, 64 and 128 clusters. For each number of clusters, we run weight clustering 3 times and report the mean values for each metric. 

\begin{figure}
  \centering
  \subfloat[Model size after clustering.]{\includegraphics[width=0.8\columnwidth]{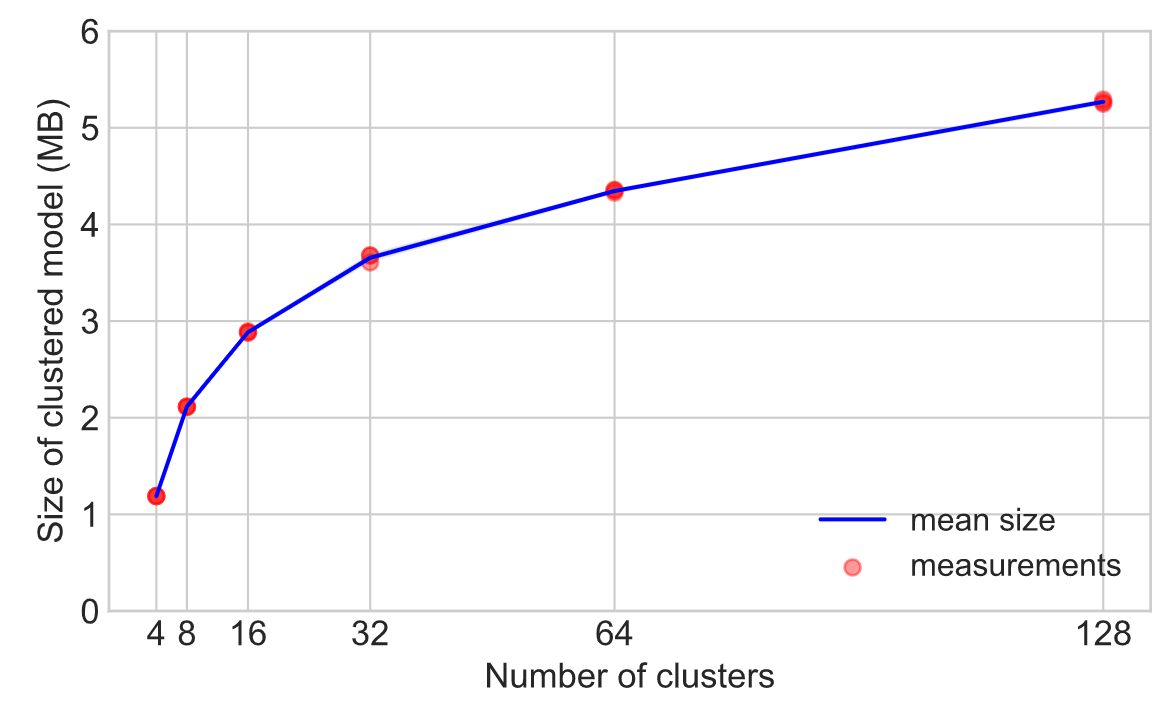}}
  \hfill
  \subfloat[Model accuracy after weight clustering. Standard deviation is shaded in gray.]{\includegraphics[width=0.8\columnwidth]{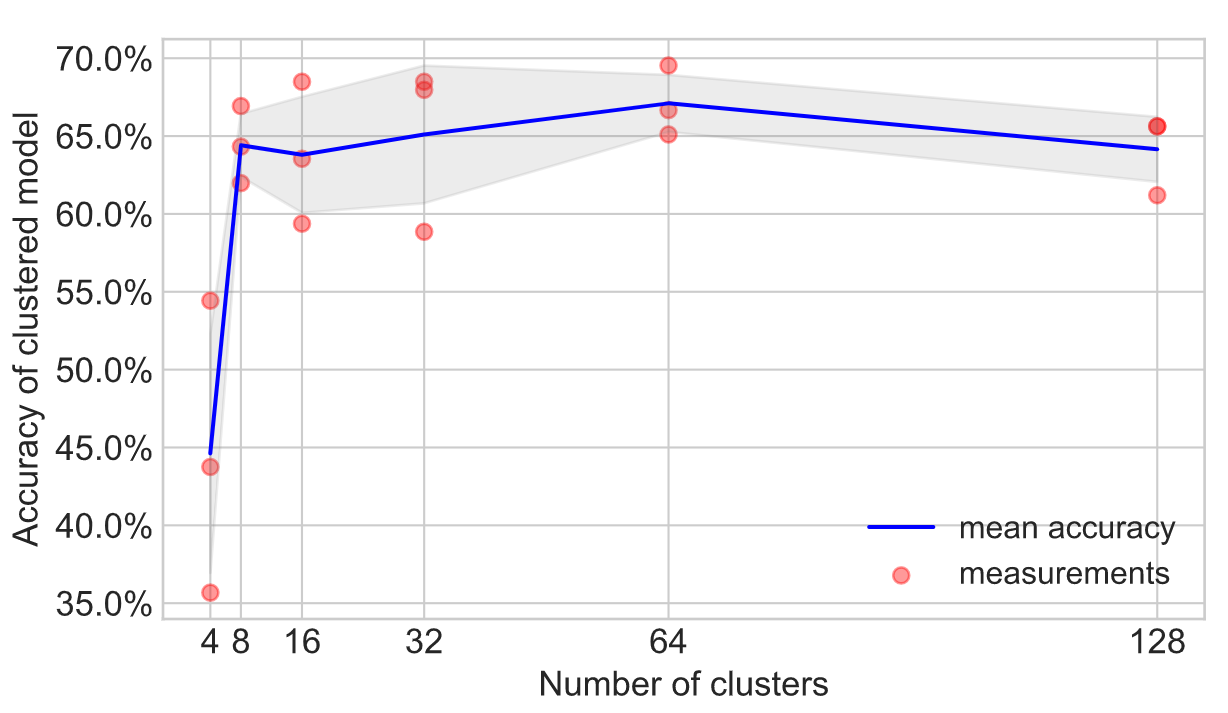}}
  \caption{Weight clustering's effect on size and overall accuracy for CK+ DB.}
  \label{fig:clustering_tradeoff}
\end{figure}

Similar to pruning, the ``number of clusters'' parameter introduces a size-accuracy trade-off illustrated in Figure \ref{fig:clustering_tradeoff}. Decreasing the number of clusters rapidly decreases the baseline model size (shrinking it by almost 14 times at 4 clusters). However, an excessively low number of clusters can decrease accuracy dramatically (with overall accuracy dropping below 45\% at 4 clusters). 

Despite that, we can see that keeping the number of clusters sufficiently high (between 8 and 128) preserves overall accuracy close to the baseline accuracy of 67.96\%, while offering a lucrative reduction in model size. 

Just like with pruning, though, the overall accuracy of the clustered models is deceptive as it hides an increasing gap between male and female accuracy. Table \ref{tab:clustering_fairness} shows that the baseline 2.36\% gap has increased massively, reaching 19.77\% in favour of male subjects at 16 clusters. 

Applying quantisation to the clustered models reduces their size by a further 17\% on average – a much smaller reduction than was observed for pruning. Again, though, quantisation comes at no cost for accuracy or fairness – overall accuracy, male accuracy and female accuracy have all stayed within 1.5\% of the values for the clustered models.

\begin{table}
  \caption{Weight clustering's effect on fairness for CK+ DB.}
  \label{tab:clustering_fairness}
  \small
  \begin{tabular}{c|cc|c}
    \toprule
    Number of clusters &Female accuracy&Male accuracy&Gap\\
    \midrule
    4 & 43.05\% & 47.22\% & 4.17\%\\
    8 & 59.30\% & 72.91\% & 13.61\%\\
    16 & 56.38\% & 76.15\% & \textbf{19.77}\%\\
    32 & 60.55\% & 72.68\% & 12.13\%\\
    64 & 60.27\% & 78.47\% & 18.20\%\\
    128 & 59.16\% & 72.45\% & 13.29\%\\
  \bottomrule
\end{tabular}
\end{table}

\subsection{RAF-DB Experiments}

We now present the results obtained on the RAF-DB dataset. We provide the full results in Table \ref{tab:extended_results_rafdb} in the appendix, and only summarise the more interesting results below. 

\subsubsection{Baseline Performance}

The performance of the baseline RAF-DB classifier is summarised in Table \ref{tab:rafdb_baseline_performance}. The baseline model has a size of \textbf{29.80 MB} and reports overall test accuracy of \textbf{82.46\%}, which seems acceptable given that much larger, state-of-the-art architectures have reported between 86\% and 89\% on this dataset \cite{rafdb_state_of_the_art}. 
In terms of fairness, the classifier seems to classify female subjects slightly more accurately than male subjects, and African-American subjects better than Caucasian or Asian subjects. However, those differences are minor. A more major classification gap is observed with respect to the age attribute where there is a 19\% classification gap between the best classified age group (A0 or 0-3 years old) and the worst classified age group (A4 or 70+ years old). This could be due to younger subjects expressing emotions more explicitly, or a different distribution of emotion labels across the two age groups. Such a conclusion is supported by relevant works indicating that the age of the face plays an important role for facial expression decoding and factors such as lower expressivity and age-related changes in the face may lower decoding accuracy for older faces \cite{FolsterEtAl-2014}.

\begin{table}
  \caption{Metrics for the baseline RAF-DB classifier.} 
  \label{tab:rafdb_baseline_performance}
  \small
  \begin{tabular}{c|c}
    \toprule
    \textbf{Metric}&\textbf{Value}\\
    \midrule
    \midrule
    Size & 29.80 MB\\
    Overall accuracy & 82.46\%\\
    \midrule
    Female accuracy & 83.33\%\\
    Male accuracy & 80.54\%\\
    \midrule
    Caucasian accuracy & 81.92\%\\
    African-American accuracy & 86.75\%\\
    Asian accuracy & 83.02\%\\
    \midrule
    A0 accuracy & 89.96\%\\
    A1 accuracy & 82.96\%\\
    A2 accuracy & 80.44\%\\
    A3 accuracy & 85.85\%\\
    A4 accuracy & 70.78\%\\
  \bottomrule
\end{tabular}
\end{table}

\subsubsection{Quantisation Results}

\begin{figure}
  \centering
  \includegraphics[width=0.8\columnwidth]{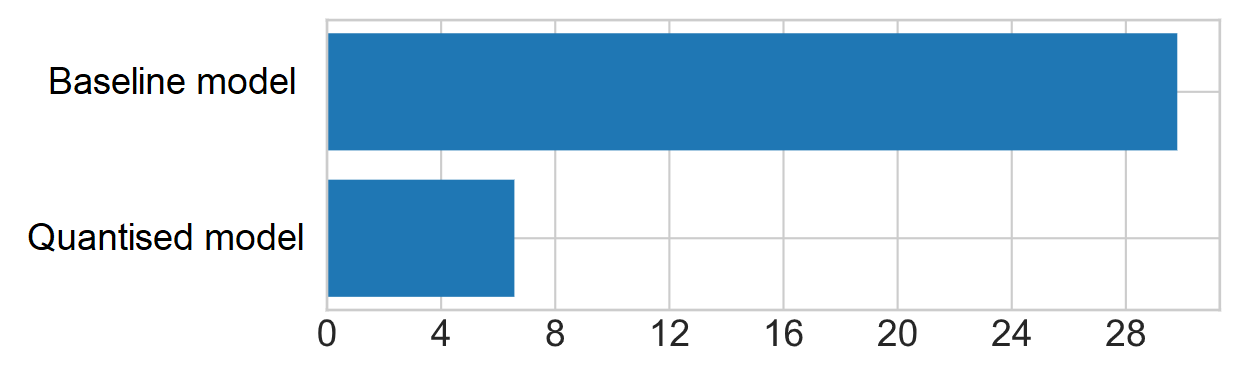}
  \caption{Size of the RAF-DB classifier before (top) and after (bottom) applying quantisation.}
  \label{fig:quantisationRAF}
\end{figure}

Figure~\ref{fig:quantisationRAF} shows the size of the RAF-DB classifier before (top) and after (bottom) applying quantisation. We observe that applying quantisation to the baseline classifier reduces the size of the model by around 4.5 times - from 29.80 MB down to 6.56 MB. At the same time, similar to the CK+ experiments, applying quantisation does not impact the predictive performance of the classifier. The quantised model's overall accuracy is identical to the one reported by the baseline model (82.46\%), and Table \ref{tab:extended_results_rafdb} shows that none of the per-class accuracies have changed by more than 0.2\%. 

\subsubsection{Pruning Results}

\begin{figure}
  \centering
  \subfloat[Model size after pruning.]{\includegraphics[width=0.4\columnwidth]{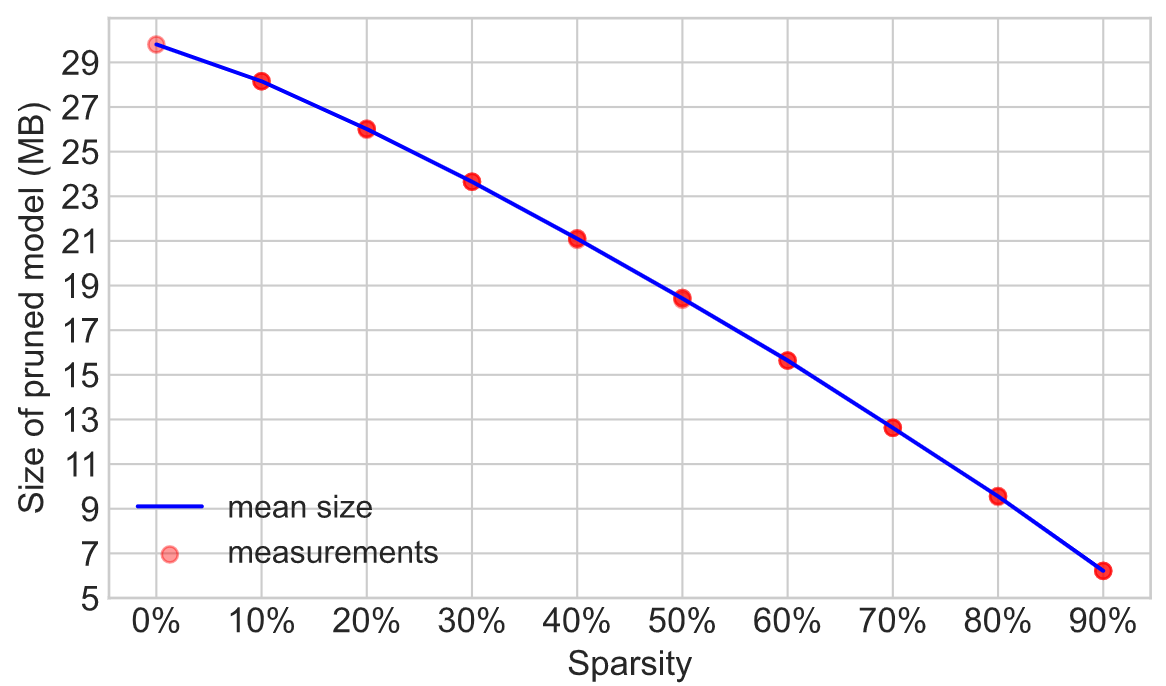}}
  \hfill
  \subfloat[Model accuracy after pruning. Standard deviation is shaded in gray.]{\includegraphics[width=0.4\columnwidth]{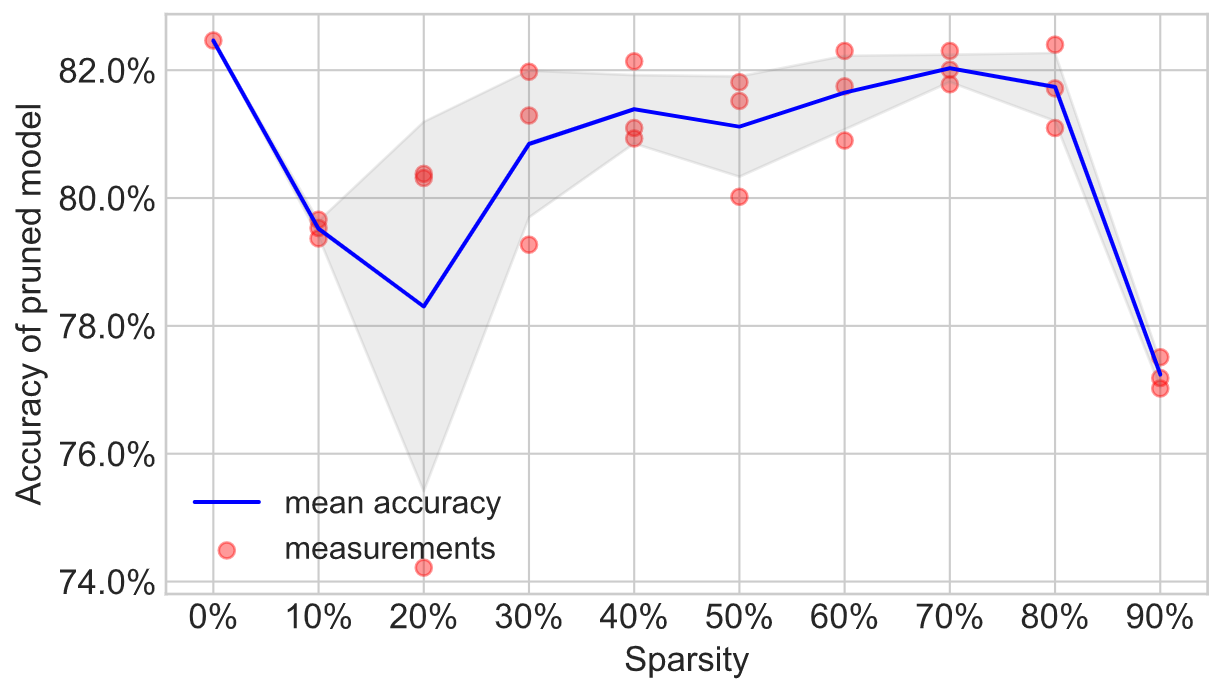}}
  \caption{Pruning's effect on model size and accuracy for RAF-DB.}
  \label{fig:rafdb_pruning_tradeoff}
\end{figure}

Similar to the CK+ experiments, we observe that applying pruning to the RAF-DB classifier translates to a linear reduction in model size as illustrated in Figure \ref{fig:rafdb_pruning_tradeoff}. Unlike the CK+ classifier though, the accuracy of the RAF-DB model seems to be much more robust to pruning. Even at 80\% sparsity, the overall accuracy has only dropped down to 81.73\% compared to the baseline (82.46\%). Only when sparsity increases to 90\% do we observe a more significant drop in accuracy down to 77.23\%. The RAF-DB model is therefore more akin to models such as MNIST classifiers \cite{mnistSparsity} where near-optimal accuracy can be preserved even at 99\% sparsity. We analyse the difference in robustness between the CK+ and the RAF-DB model, and discuss potential causes for this disparity in the Discussion section. 

As for gender fairness, the original 2.8\% classification gap varies between 0.2\% and 2.7\% depending on the level of sparsity, so there is no evidence suggesting that sparsity has negatively impacted fariness. Since RAF-DB also has race and age labels, we can analyse fairness across those dimensions as well. The full fairness metrics are presented in Table \ref{tab:extended_results_rafdb}, which shows that applying pruning to the RAF-DB classifier has little to no effect on model age and race fairness too. 

\subsubsection{Weight Clustering Results}
 Figure \ref{fig:rafdb_pruning_tradeoff} shows pruning's effect on model size and accuracy for RAF-DB. In terms of model accuracy, the classifier trained and tested on RAF-DB seems more robust to compression compared to the CK+ experiments. Regardless of the level of sparsity or the number of clusters of the compression strategy, the overall accuracy of the compressed model does not fall below 77\%, which is not significantly lower than the uncompressed (or baseline) model, which has an accuracy of 82\%. 


\begin{figure}
  \centering
  \subfloat[Model size after clustering.]{\includegraphics[width=0.8\columnwidth]{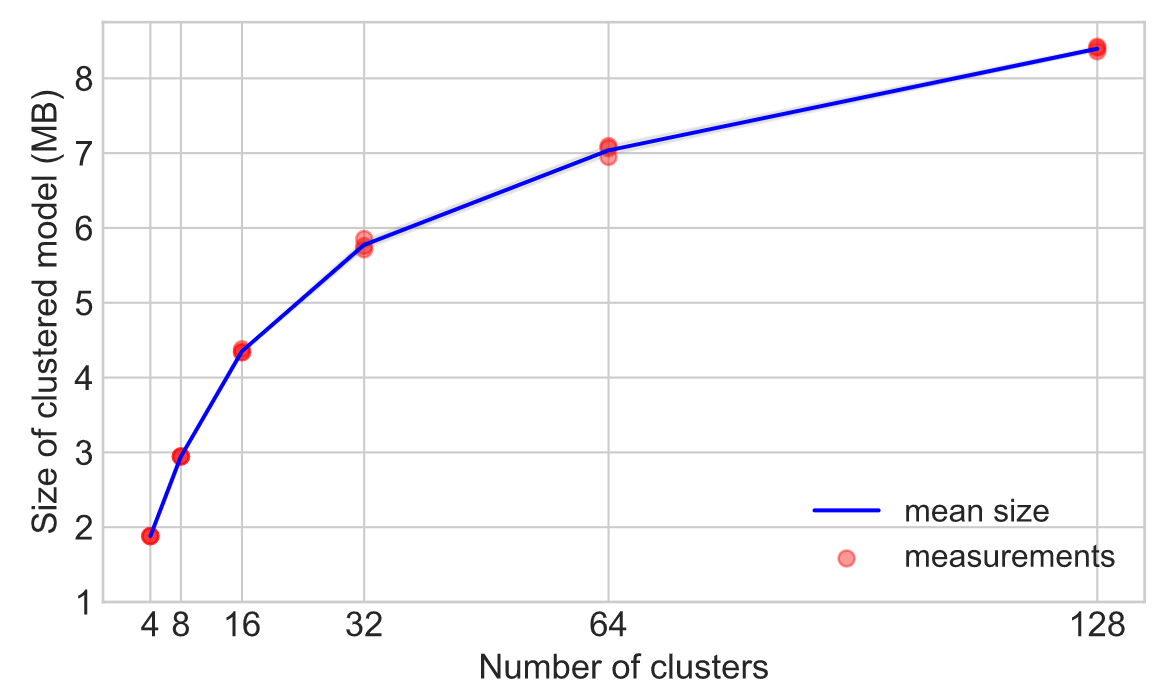}}
  \hfill
  \subfloat[Model accuracy after weight clustering. Standard deviation is shaded in gray.]{\includegraphics[width=0.8\columnwidth]{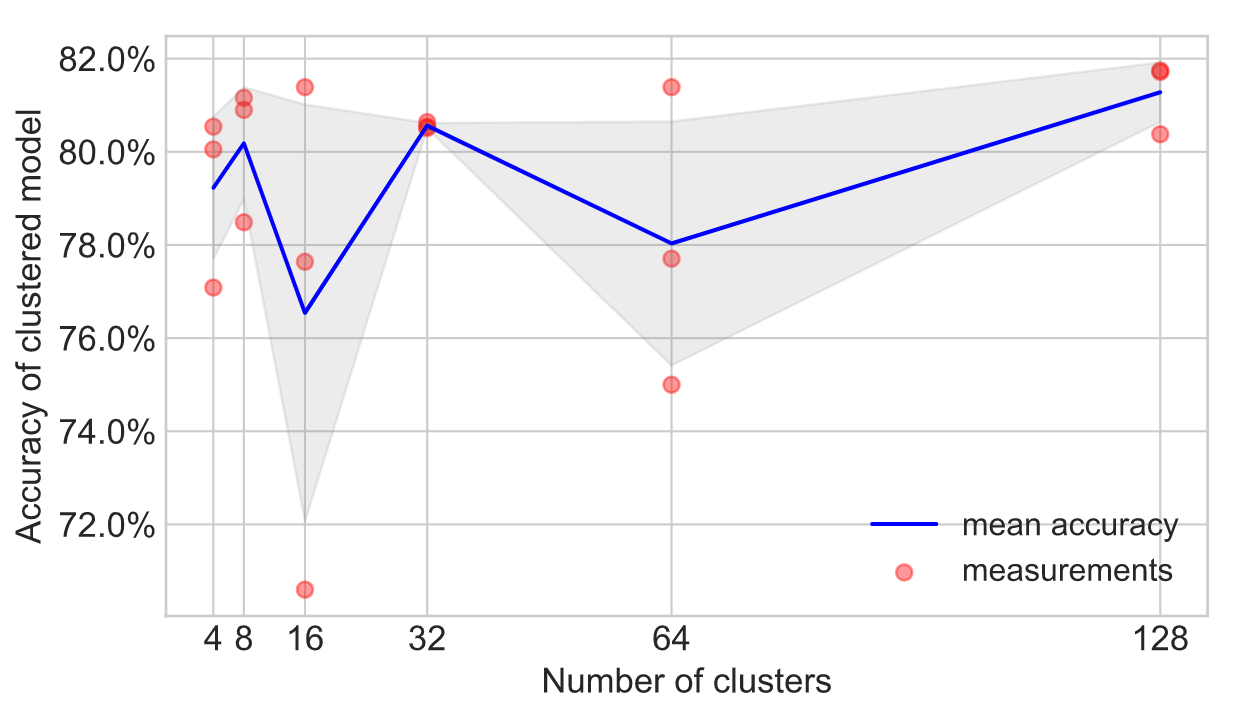}}
  \caption{Weight clustering's effect on size and overall accuracy for RAF DB.}
  \label{fig:clustering_tradeoff_RAF}
\end{figure}

\section{Discussion} 

\begin{figure}
  \centering
  \subfloat{\includegraphics[width=0.50\columnwidth]{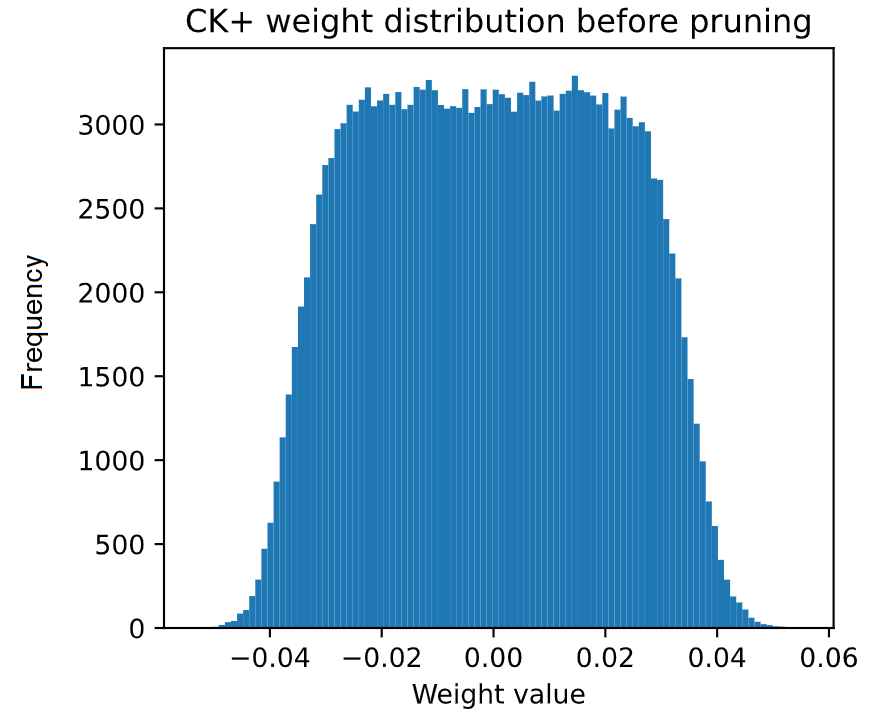}}
  \hfill
  \subfloat{\includegraphics[width=0.50\columnwidth]{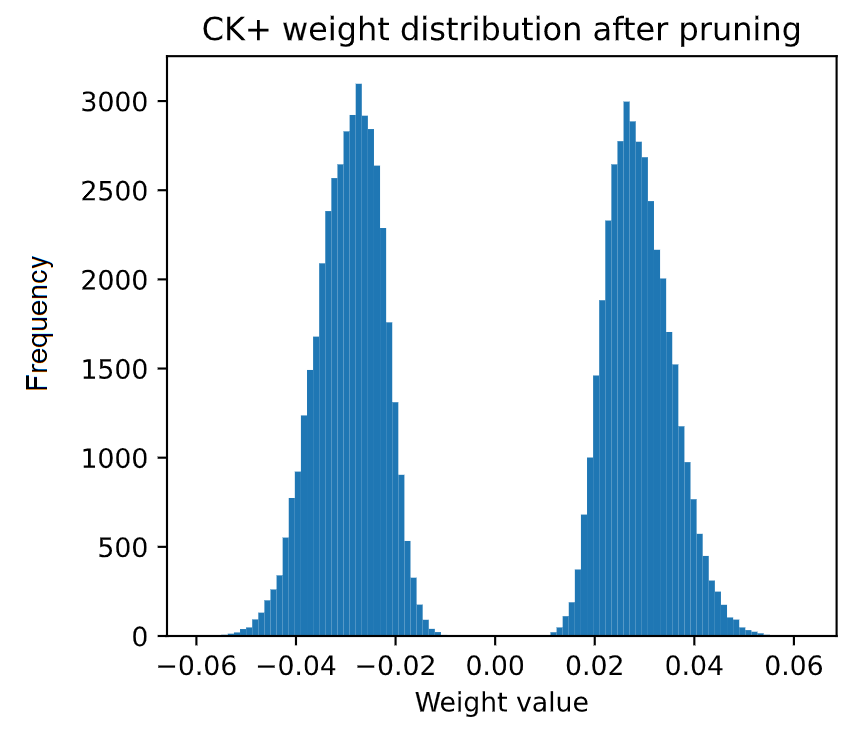}}
  \caption{Distribution of the kernel weights of the \texttt{conv2d} layer of the CK+ classifier before (left) and after (right) pruning at 60\%.}
  \label{fig:ckplus_weight_distribution}
\end{figure}

\begin{figure}
  \centering
  \subfloat{\includegraphics[width=0.50\columnwidth]{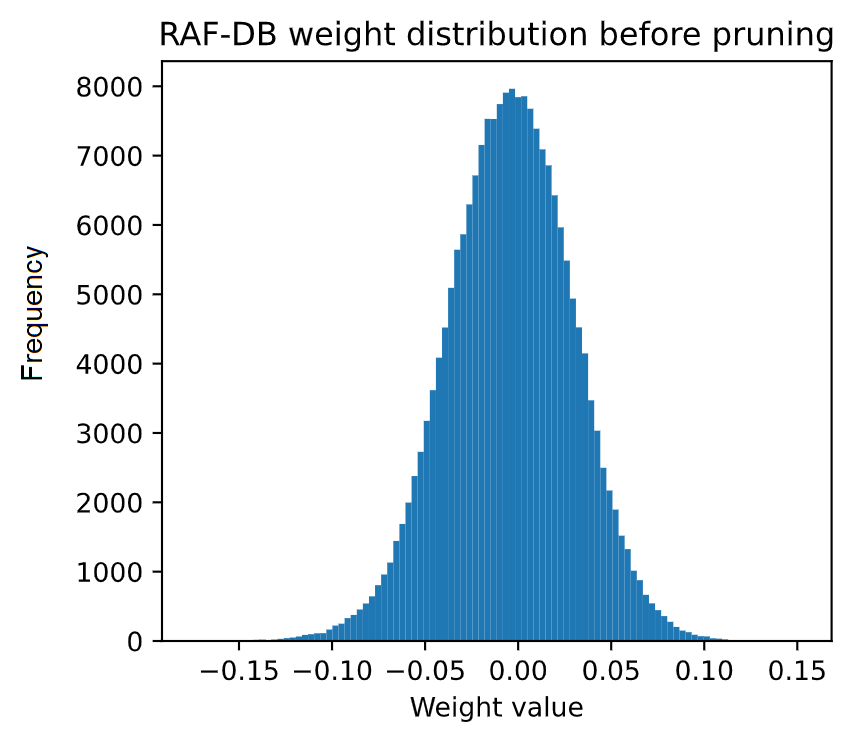}}
  \hfill
  \subfloat{\includegraphics[width=0.50\columnwidth]{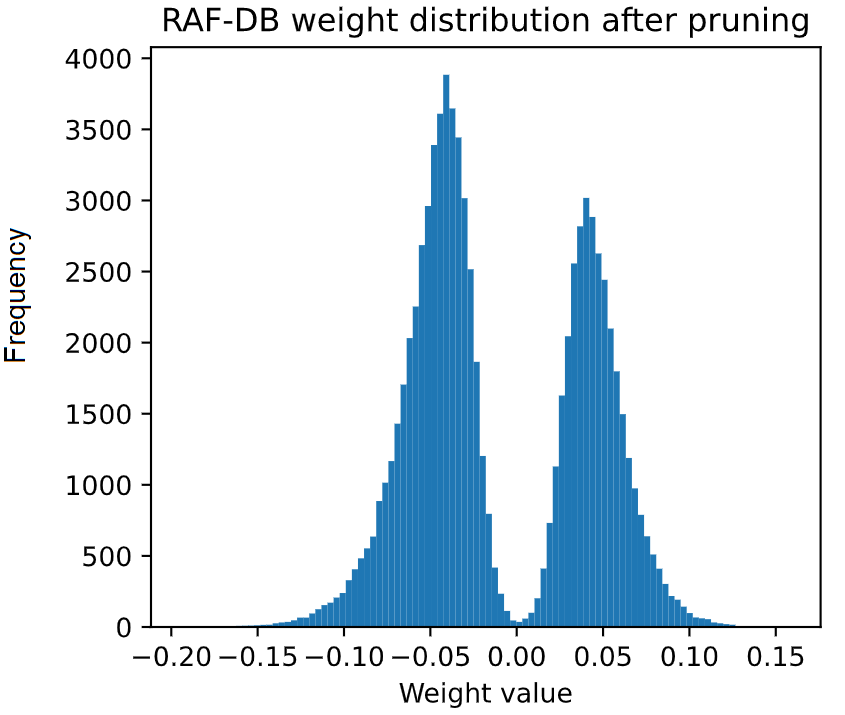}}
  \caption{Distribution of the kernel weights of the \texttt{conv2d} layer of the RAF-DB classifier before (left) and after (right) pruning at 60\%.}
  \label{fig:rafdb_weight_distribution}
\end{figure}

In this study, we analysed and compared the effect of model compression on model size, accuracy and fairness in the context of facial expression recognition on two facial expression datasets. We now revisit our research questions from Section \ref{project_goals}:

\begin{itemize}
    \item \textbf{RQ1: \textit{“How effective is model compression in the context of FER?”}} We saw that model compression can dramatically reduce the storage requirements of both FER models. Quantisation alone achieves around $4-4.5 \times$ reduction in model sizes with minimal impact on overall accuracy for both CK+ and RAF-DB. Both pruning and clustering progressively decrease the model size on disk as sparsity increases (and the number of clusters decrease).
    In terms of model accuracy, the classifier trained and tested on RAF-DB seems more robust to compression compared to the CK+ DB one. Regardless of the level of sparsity or the number of clusters of the compression strategy, the overall accuracy of the compressed model does not fall below 77\%. 
    \item \textbf{RQ2: \textit{“Do model compression techniques amplify biases?”}} Our findings for CK+ DB confirm the claims by previous studies that model compression can amplify existing biases for gender for deep learning models trained for facial expression recognition. However our findings for RAF-DB indicate that sparsity does not impact fairness in terms of gender, race or age negatively. 
    \item \textbf{RQ3: \textit{“Is the impact on fairness identical across different compression techniques?”}} Our results for CK+ DB suggest that different compression techniques tend to have a highly distinct impact on fairness: Post-training quantisation has no visible effect on baseline fairness. Meanwhile, pruning and weight clustering can severely amplify biases, increasing the initial 2.36\% gap between male and female accuracy to 15.60\% and 19.77\% respectively. On the other hand for RAF-DB we find that
    the different compression strategies do not seem to increase the gap in predictive performance across any of the three demographic attributes (gender, race and age). That is in contrast with the CK+ experiment findings where compression seems to amplify existing biases. 
\end{itemize}

In order to understand the reasons for the different findings related to these datasets, we compare the weight distributions of CK+ and RAF-DB before pruning, and we see that the CK+ distribution has a `wider' shape compared to the RAF-DB one which is much more `narrow' and most of its values are located close to its mean at 0.00. As a result, when we prune the two models, we get gaps of different sizes. For the `wide' CK+ distribution, when we prune at 60\%, we need to set `crop' or zero-out 60\% of its weights. However, the CK+ weights are relatively evenly distributed and most of them are located at some (relative) distance from the 0.00 mean. As a result, we need to `crop' weights which are not located immediately around the centre and that causes a wide gap in the middle. We can expect that a large gap will have a bigger impact on predictive performance since it means values which are further from 0.00 have been set to 0.00. Meanwhile, for the RAF-DB the resulting gap is much smaller. This is because the original distribution is much more `narrow' in terms of shape with a big share of the weights located at or close to 0.00. Therefore, when we prune at 60\%, the 60\% of the weights which we set to 0.00 are going to be already equal to or close to 0.00 and therefore we can expect a minor impact on predictive performance.
This can explain why RAF-DB is much more robust than CK+ to pruning. As a reminder, at 60\% pruning, CK+’s
accuracy dropped from 68\% down to 42\%, and RAF-DB’s accuracy only dropped from 82.4\% down to 81.6\%.
While the plots in this paper only show the weights from the first dense layer, we observe a similar trend for the weights of the other dense layers, as well as the weights of the convolutional layers. It is important to note that the CK+DB is much more homogenous in terms of acquisition setup and setting as compared to RAF-DB which contains various facial images crawled from the internet, and the size of CK+ DB is significantly smaller than RAF-DB.



\section{Limitations and Future Work}

We identify a couple of \textbf{limitations} of our study: First, the gender annotation of the CK+ dataset was performed manually. While the annotations were straightforward, labels could be obtained via a crowdworking experiment or a user study to better ensure the ground truth is reliable and unbiased. 

Furthermore, our baseline model is not a state-of-the-art FER classifier as of 2021. We deliberately selected a relatively small architecture that could be trained, fine-tuned and evaluated locally, but given more time and computational resources the study could be extended to consider a larger and more modern architecture. 

The work in this paper could be extended in several different directions: More compression strategies (e.g., quantisation-aware training and various forms of weight sharing \cite{weightvirt}) could also be evaluated. The study could also be extended to 
explore more system metrics such as latency, memory consumption, etc. 

\section{Acknowledgments}
S. Stoychev completed this work while studying towards his MPhil in Advanced Computing in the Department of Computer Science and Technology, University of Cambridge.
H. Gunes is supported by the EPSRC project ARoEQ under grant ref. EP/R$030782$/$1$. 

\bibliographystyle{ACM-Reference-Format}
\bibliography{sample-base}

\appendix

\section{Full CK+ Results}

\begin{table}[H]
  \caption{Full results from the CK+ experiments. }
  \label{tab:extended_results}
  \small
  \begin{tabular}{c|cccc}
    \toprule
    Model&Size (MB) &Overall acc.&Female acc.&Male acc.\\
    \midrule
    baseline & 16.51 & 67.96\% & 67.08\% & 69.44\%\\
    \midrule 
    quantised & 4.04 & 67.96\% & 67.08\% & 69.44\%\\
    \midrule 
    pruned (10\%) & 15.56 & 65.71\% & 59.86\% & 75.46\%\\
    pruned (20\%) & 14.31 & 67.62\% & 62.77\% & 75.69\%\\
    pruned (30\%) & 12.97 & 67.36\% & 62.36\% & 75.69\%\\
    pruned (40\%) & 11.55 & 65.79\% & 63.47\% & 69.67\%\\
    pruned (50\%) & 10.10 & 64.06\% & 59.86\% & 71.06\%\\
    pruned (60\%) & 8.58 & 41.11\% & 44.02\% & 36.34\%\\
    \midrule
    
    pruned (10\%) + quant. & 3.88 & 65.45\% & 59.58\% & 75.23\%\\
    pruned (20\%) + quant. & 3.73 & 67.53\% & 62.50\% & 75.92\%\\
    pruned (30\%) + quant. & 3.51 & 67.36\% & 62.36\% & 75.69\%\\
    pruned (40\%) + quant. & 3.21 & 66.05\% & 63.88\% & 69.67\%\\
    pruned (50\%) + quant. & 2.85 & 64.06\% & 59.86\% & 71.06\%\\
    pruned (60\%) + quant. & 2.44 & 40.88\% & 43.61\% & 36.34\%\\
    \midrule
    
    clustered (4 cl.) & 1.18 & 44.61\% & 43.05\% & 47.22\%\\
    clustered (8 cl.) & 2.11 & 64.40\% & 59.30\% & 72.91\%\\
    clustered (16 cl.) & 2.88 & 63.80\% & 56.38\% & 76.15\%\\
    clustered (32 cl.) & 3.65 & 65.10\% & 60.55\% & 72.68\%\\
    clustered (64 cl.) & 4.34 & 67.10\% & 60.27\% & 78.47\%\\
    clustered (128 cl.) & 5.26 & 64.14\% & 59.16\% & 72.45\%\\
    \midrule
    
    clust. (4 cl.) + quant. & 0.91 & 43.57\% & 42.08\% & 46.06\%\\
    clust. (8 cl.) + quant. & 1.81 & 64.49\% & 59.72\% & 72.45\%\\
    clust. (16 cl.) + quant. & 2.57 & 64.49\% & 57.08\% & 76.85\%\\
    clust. (32 cl.) + quant. & 3.22 & 65.71\% & 61.25\% & 73.14\%\\
    clust. (64 cl.) + quant. & 3.63 & 66.75\% & 60.00\% & 78.00\%\\
    clust. (128 cl.) + quant. & 3.79 & 64.49\% & 59.72\% & 72.45\%\\
    
  \bottomrule
\end{tabular}
\end{table}









\onecolumn
\section{Full RAF-DB Results}

\begin{table}[H]
  \caption{Full results from the RAF-DB experiments. \textit{``Female''} and \textit{``Male''} indicate accuracies for female and male subjects respectively. \textit{``Cauc.''}, \textit{``Af.-Am.''} and \textit{``Asian''} denote accuracies for subjects labelled as Caucasian, African-American and Asian respectively. \textit{``A0''} to \textit{``A4''} indicate accuracies across the 5 age groups - 0-3, 4-19, 20-39, 40-69 and 70+.} 
  \label{tab:extended_results_rafdb}
  \small
  \begin{tabular}{c|cc|cc|ccc|ccccc}
    \toprule
    Model&Size (MB) & Overall acc. & Female & Male & Cauc. & Af.-Am. & Asian & A0 & A1 & A2 & A3 & A4 \\
    \midrule
    baseline & 29.80 & 82.46\% & 83.33\% & 80.54\% & 81.92\% & 86.75\% &  83.02\% & 89.96\% & 82.92\% & 80.44\% & 85.85\% & 70.78\% \\
    \midrule 
    quantised & 6.56 & 82.46\% & 83.20\% & 80.70\% & 81.92\% & 86.75\% &  83.02\% & 89.96\% & 82.92\% & 80.50\% & 85.65\% & 70.78\% \\
    \midrule 
    pruned (10\%) & 28.15 & 79.51\% & 80.04\% & 78.64\% & 79.00\% & 81.90\% & 80.88\% & 82.57\% & 82.64\% & 78.53\% & 79.88\% & 67.41\% \\
    pruned (20\%) & 26.01 & 78.30\% & 78.72\% & 77.47\% & 78.10\% & 80.19\% & 78.32\% & 82.16\% & 79.35\% & 77.47\% & 79.34\% & 67.79\% \\
    pruned (30\%) & 23.65 & 80.84\% & 81.37\% & 79.69\% & 80.44\% & 84.33\% & 81.09\% & 85.51\% & 82.71\% & 78.90\% & 84.32\% & 70.03\% \\
    pruned (40\%) & 21.09 & 81.38\% & 81.95\% & 79.93\% & 80.97\% & 84.04\% & 82.10\% & 88.34\% & 82.78\% & 79.56\% & 83.79\% & 68.53\% \\
    pruned (50\%) & 18.41 & 81.11\% & 81.83\% & 79.47\% & 80.85\% & 84.33\% & 80.81\% & 87.03\% & 82.85\% & 79.18\% & 83.59\% & 71.91\% \\
    pruned (60\%) & 15.63 & 81.64\% & 82.53\% & 79.82\% & 81.18\% & 83.76\% & 82.88\% & 87.03\% & 82.57\% & 80.42\% & 82.73\% & 73.40\% \\
    pruned (70\%) & 12.62 & 82.02\% & 82.40\% & 80.62\% & 81.75\% & 84.90\% & 81.98\% & 87.84\% & 83.88\% & 80.22\% & 84.32\% & 71.16\% \\
    pruned (80\%) & 9.55 & 81.73\% & 82.26\% & 80.11\% & 81.35\% & 84.90\% & 82.05\% & 88.34\% & 83.19\% & 79.88\% & 83.86\% & 71.91\% \\
    pruned (90\%) & 6.21 & 77.23\% & 77.09\% & 76.83\% & 76.77\% & 78.20\% & 79.02\% & 82.26\% & 78.87\% & 75.35\% & 80.34\% & 67.41\% \\
    \midrule
    
    pruned (10\%) + quant. & 6.44 & 79.54\% & 80.06\% & 78.62\% & 79.05\% & 81.90\% & 80.74\% & 82.67\% & 82.57\% & 78.58\% & 79.94\% & 67.04\% \\
    pruned (20\%) + quant. & 6.29 & 78.32\% & 78.68\% & 77.58\% & 78.16\% & 80.19\% & 78.19\% & 82.06\% & 79.49\% & 77.45\% & 79.54\% & 67.41\% \\
    pruned (30\%) + quant. & 6.03 & 80.85\% & 81.44\% & 79.58\% & 80.50\% & 84.33\% & 80.88\% & 85.71\% & 82.78\% & 78.92\% & 84.19\% & 69.66\% \\
    pruned (40\%) + quant. & 5.55 & 81.46\% & 82.05\% & 79.98\% & 81.05\% & 84.04\% & 82.19\% & 88.34\% & 82.78\% & 79.68\% & 83.86\% & 68.53\% \\
    pruned (50\%) + quant. & 5.02 & 81.04\% & 81.83\% & 79.47\% & 80.78\% & 84.33\% & 80.67\% & 87.03\% & 82.71\% & 79.18\% & 83.20\% & 72.28\% \\
    pruned (60\%) + quant. & 4.38 & 81.64\% & 82.46\% & 79.90\% & 81.19\% & 83.76\% & 82.81\% & 86.93\% & 82.51\% & 80.42\% & 82.86\% & 73.40\% \\
    pruned (70\%) + quant. & 3.57 & 82.05\% & 82.42\% & 80.65\% & 81.79\% & 84.75\% & 81.98\% & 87.84\% & 84.01\% & 80.26\% & 84.19\% & 71.16\% \\
    pruned (80\%) + quant. & 2.73 & 81.70\% & 82.26\% & 80.11\% & 81.36\% & 84.75\% & 81.84\% & 88.14\% & 83.19\% & 79.76\% & 84.19\% & 71.91\% \\
    pruned (90\%) + quant. & 1.65 & 77.33\% & 77.26\% & 76.88\% & 76.88\% & 77.92\% & 79.22\% & 82.16\% & 78.94\% & 75.49\% & 80.47\% & 67.41\% \\
    \midrule
    
    clustered (4 cl.) & 1.88 & 79.22\% & 79.71\% & 78.08\% & 79.05\% & 79.05\% & 80.12\% & 85.61\% & 81.55\% & 76.93\% & 81.80\% & 71.16\% \\
    clustered (8 cl.) & 2.94 & 80.18\% & 80.39\% & 79.34\% & 79.75\% & 81.48\% & 81.64\% & 85.10\% & 81.61\% & 78.68\% & 82.27\% & 70.41\% \\
    clustered (16 cl.) & 4.35 & 76.54\% & 77.44\% & 75.02\% & 75.92\% & 78.77\% & 78.46\% & 80.64\% & 79.21\% & 75.15\% & 78.55\% & 61.42\% \\
    clustered (32 cl.) & 5.77 & 80.56\% & 80.94\% & 79.42\% & 80.13\% & 82.62\% & 81.64\% & 86.32\% & 82.30\% & 78.90\% & 82.60\% & 69.28\% \\
    clustered (64 cl.) & 7.03 & 78.03\% & 78.00\% & 77.55\% & 77.38\% & 81.48\% & 79.50\% & 82.97\% & 79.90\% & 76.07\% & 80.94\% & 69.66\% \\
    clustered (128 cl.) & 8.39 & 81.27\% & 81.83\% & 80.19\% & 80.77\% & 82.62\% & 83.09\% & 86.72\% & 82.85\% & 79.82\% & 83.06\% & 69.66\% \\
    \midrule
    
    clust. (4 cl.) + quant. & 1.39 & 77.89\% & 78.76\% & 76.32\% & 77.62\% & 78.49\% & 78.88\% & 82.57\% & 79.56\% & 76.11\% & 81.07\% & 66.66\% \\
    clust. (8 cl.) + quant. & 2.29 & 79.97\% & 80.02\% & 79.37\% & 79.46\% & 82.33\% & 81.29\% & 84.80\% & 81.41\% & 78.49\% & 82.33\% & 68.53\% \\
    clust. (16 cl.) + quant. & 3.53 & 76.44\% & 77.22\% & 74.99\% & 75.86\% & 78.34\% & 78.32\% & 80.54\% & 79.35\% & 75.23\% & 77.49\% & 62.17\% \\
    clust. (32 cl.) + quant. & 4.71 & 80.62\% & 81.13\% & 79.47\% & 80.15\% & 83.04\% & 81.78\% & 85.71\% & 82.57\% & 79.30\% & 81.80\% & 69.28\% \\
    clust. (64 cl.) + quant. & 5.52 & 78.00\% & 78.02\% & 77.52\% & 77.37\% & 80.62\% & 79.84\% & 82.57\% & 79.83\% & 76.15\% & 81.07\% & 68.53\% \\
    clust. (128 cl.) + quant. & 5.89 & 81.13\% & 81.68\% & 80.09\% & 80.74\% & 82.19\% & 82.53\% & 86.01\% & 82.71\% & 79.76\% & 83.20\% & 68.53\% \\
    
  \bottomrule
\end{tabular}
\end{table}

\end{document}